\documentclass[11pt,a4paper]{article}
\usepackage[hyperref]{acl2021}
\usepackage{times}
\usepackage{latexsym}

\usepackage{amsthm,amsmath,amssymb}
\usepackage{mathrsfs}
\usepackage{CJKutf8}
\usepackage{graphicx}
\usepackage{algorithm}
\usepackage{algorithmic}
\usepackage{float}
\usepackage[T2A,T1]{fontenc}
\usepackage[russian,english]{babel}
\usepackage{amssymb }
\usepackage{subfigure}
\usepackage{booktabs}

\usepackage{multirow}

\usepackage{url}
\usepackage{hyperref}

\usepackage{microtype}

\aclfinalcopy 
\def\aclpaperid{2347} 

\title{Combining Static Word Embeddings and Contextual Representations for Bilingual Lexicon Induction}

\author{
	{Jinpeng Zhang\textsuperscript{1}, Baijun Ji\textsuperscript{1}, Nini Xiao\textsuperscript{1}, Xiangyu Duan\textsuperscript{1}\thanks{  $\quad $ Corresponding Author. }, Min Zhang\textsuperscript{1} } \\
	{\textbf{Yangbin Shi\textsuperscript{2}, Weihua Luo\textsuperscript{2} } }
	\vspace{1.6mm}\\
	\fontsize{12}{10}\selectfont
	\,\textsuperscript{\rm 1}  Institute of Aritificial Intelligence, School of Computer Science and Technology, \\
	\fontsize{12}{10}\selectfont Soochow university  \\
            \fontsize{12}{10}\selectfont  \textsuperscript{\rm 2} Alibaba DAMO Academy \\
            \fontsize{10}{10}\selectfont \{jpzhang1,bjji,nnxiaoxiao\}@stu.suda.edu.cn; \{xiangyuduan,minzhang\}@suda.edu.cn;\\
	\fontsize{10}{10}\selectfont taiwu.syb@taobao.com; weihua.luowh@alibaba-inc.com\\	} 

\date{}

\begin{document}
\maketitle
\begin{abstract}
Bilingual Lexicon Induction (BLI) aims to map words in one language to their translations in another, and is typically through learning linear projections to align monolingual word representation spaces. Two classes of word representations have been explored for BLI: static word embeddings and contextual representations, but there is no studies to combine both. In this paper, we propose a simple yet effective mechanism to combine the static word embeddings and the contextual representations to utilize the advantages of both paradigms. We test the combination mechanism on various language pairs under the supervised and unsupervised BLI benchmark settings. Experiments show that our mechanism consistently improves performances over robust BLI baselines on all language pairs by averagely improving 3.2 points in the supervised setting, and 3.1 points in the unsupervised setting\footnote{Code is released at https://github.com/zjpbinary/CSCBLI}.
\end{abstract}

\section{Introduction}

Bilingual Lexicon Induction (BLI) aims to find bilingual translation lexicons from monolingual corpora in two languages \citep{haghighi2008learning,xing2015normalized,zhang2017adversarial,artetxe2017learning,conneau2017word}, and is applied on numerous NLP tasks such as POS tagging \citep{gaddy2016ten}, parsing \cite{xiao2014distributed}, and machine translation \cite{irvine2013combining,qi2018and}.

Most work on BLI learns a mapping between two static word embedding spaces, which are pre-trained on large monolingual corpora \cite{ruder2019survey}. Both linear mapping \cite{mikolov2013exploiting,xing2015normalized,artetxe2016learning,smith2017offline} and non-linear mapping \cite{mohiuddin2020lnmap} methods have been studied to align the two spaces. Recently, other than the static word embeddings, contextual representations are used for BLI due to the significant progresses on cross-lingual applications \cite{aldarmaki2019context,schuster2019cross}. Although the static word embeddings and the contextual representations exhibit properties suited for alignment, there is no works to combine the two paradigms. 

On one hand, the static word embeddings have been widely used for BLI, but one specific embedding mapping function does not ensure that in all conditions, words in a translation pair are nearest neighbors in the mapped common space. On the other hand, the contextual representations contain rich semantic information beneficial for alignment, but the dynamic contexts of word tokens pose a challenge for aligning word types. 

In this paper, we propose a combination mechanism to utilize the static word embeddings and the contextual representations simultaneously. The combination mechanism consists of two parts. The first part is the unified word representations, in which a spring network is proposed to use the contextual representations to pull the static word embeddings to better positions in the unified space for easy alignment. The spring network and the unified word representations are trained via a contrastive loss that encourages words of a translation pair to become closer in the unified space, and penalizes words of a non-translation pair to be farther. The second part is the weighted interpolation between the words similarity in the unified word representation space and the words similarity in the contextual representation space. 

We test the proposed combination mechanism in both the supervised BLI setting which can utilize a bilingual dictionary as the training set, and the unsupervised BLI setting which does not allow using any parallel resources as supervision signal. On BLI benchmark sets of multiple language pairs, our combination mechanism performs significantly better than systems using only the static word embeddings and systems using only the contextual representations. Our mechanism improves over robust BLI baselines on all language pairs, achieving average 3.2 points improvement in the supervised setting, and average 3.1 points improvement in the unsupervised setting.

\section{Background}

The early works on Bilingual Lexicon Induction (BLI) date back to several decades ago, including feature-based retrieval \citep{fung1998ir}, distributional hypothesis \citep{rapp1999automatic,vulic2013study}, and decipherment \citep{ravi2011deciphering}. Following \citet{mikolov2013exploiting}, which pioneered the embedding based BLI method, word representation based method becomes the dominant approach, and can be categorized into two classes: static word embedding based method, and contextual representation based method. 

\begin{figure*}[htbp]
\flushleft
\centerline {\includegraphics[height=7cm,width=16cm]{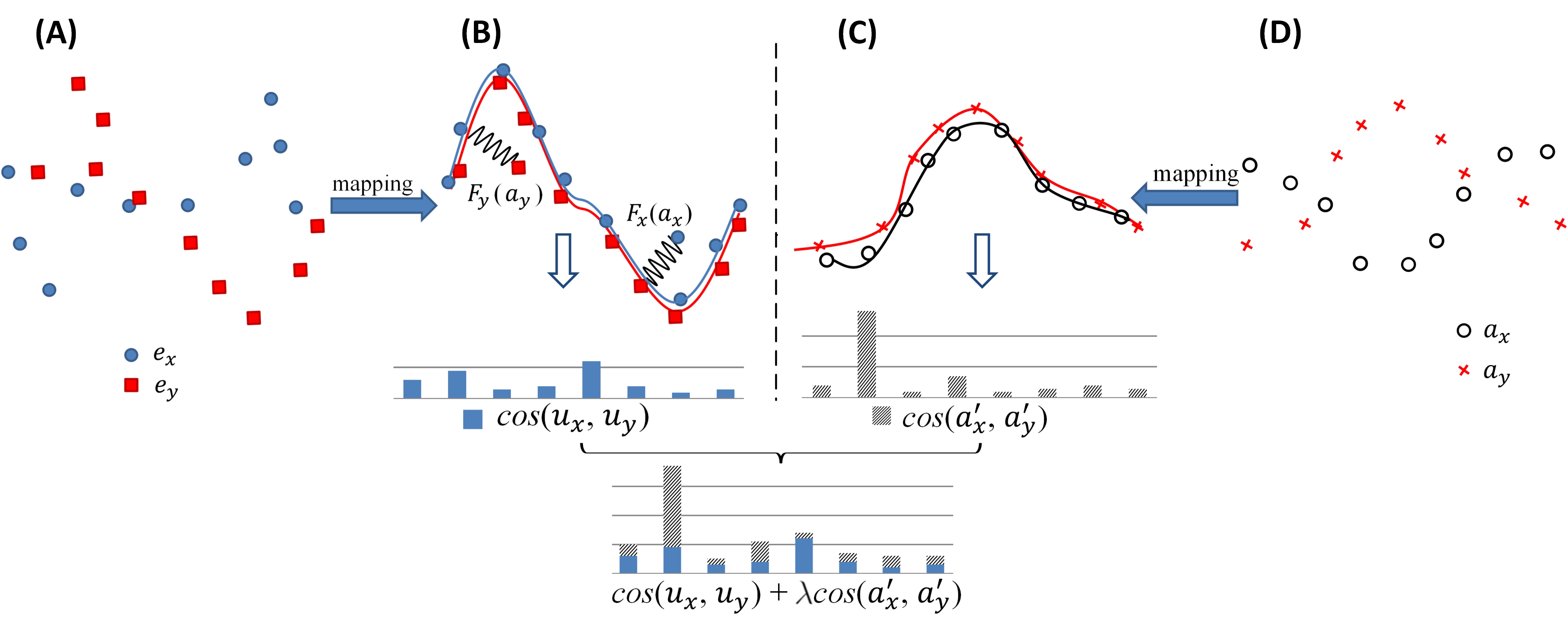} }
\caption{The illustration of the proposed combination mechanism. (A) is the static word embedding space, where $e_x$ and $e_y$ are the source and target embeddings, respectively. (B) is the unified word representation space which consists of the mapped word embeddings pulled by a spring network $F_x/F_y$ with the contextual representations as input. We just depict two springs for illustration. (C) is the mapped contextual representation space. (D) is the original contextual representation space. $a_x$ and $a_y$ are the source and target contextual representations, i.e., the average anchors, respectively. In the similarity interpolation shown in the bottom, $u_x$ and $u_y$ are the unified word representations in the two languages, $a_x^{'}$ and $a_y^{'}$ are the mapped contextual representations in the two languages, $cos$ denotes the cosine similarity function, $\mathcal{\lambda}$ denotes the weight.}
\label{fig:framework}
\end{figure*}

\begin{itemize}

\item \textbf{Static Word Embedding Based Method}

Word embeddings of different languages are pre-trained in large monolingual corpora independently. Then a mapping function is applied to align the embedding spaces of the two languages \citep{mikolov2013exploiting,xing2015normalized,artetxe2016learning,smith2017offline}. 

\hspace{3mm} We follow one robust BLI system VecMap \citep{artetxe2018aaai,artetxe2018acl}, which maps both source space and target space into a third common space. Let $E_x$ and $E_y$ be the word embedding matrices in two languages for a given bilingual dictionary such that their $i$th rows are the embeddings of words of the $i$th translation pair in the dictionary. The training objective is to find mapping functions $W_x$ and $W_y$ such that

\begin{equation}
W_x^{\star}, W_y^{\star} = \underset{W_x,W_y\in M_d(\mathbb{R})} {\mathop{\arg\max}} cos(E_x W_x, E_y W_y)    \label{equ:base}
\end{equation}

\noindent where $d$ is the dimension of the embeddings, $M_d(\mathbb{R})$ is the space of $d \times d$ matrices of real numbers. The optimal $W_x$ and $W_y$ maximizes the cosine similarity between words of each translation pair in the mapped common space. In the unsupervised version where no bilingual dictionary is given, an artificial dictionary is initialized and iteratively updated through training $W_x$ and $W_y$  according to equation (\ref{equ:base}) \citep{artetxe2018acl}.

\hspace{3mm} Both mapping functions are constrained to be orthogonal during training by setting $W_x=U$ and $W_y=V$, where $U \Sigma V^T = X^TY$ is the singular value decomposition of $X^TY$. Such orthogonal constraint is based on the assumption that the source embedding space and the target embedding space are isometric, which is a particularly strong assumption that does not hold in all conditions \citep{zhang2017earth,sogaard-etal-2018-limitations}. To depart from the isometry assumption, \citet{patra2019bilingual} uses a semi-supervised technique that leverages both seed dictionary and a larger set of unaligned word embeddings, \citet{mohiuddin2020lnmap} uses a non-linear mapping function that is not constrained to be orthogonal.

\hspace{3mm} We propose another method to relax the isometry assumption by combining the contextual representations with the word embeddings to compensate the shortage of the overly strong assumption.

\item \textbf{Contextual Representation Based Method}

Contextual representations can be obtained through multilingual pre-training, which encodes whole sentence and outputs contextual representation for each word \citep{devlin-etal-2019-bert,lample2019cross}. Due to the rich context information contained in the contextual representations, there are endeavors to align them in different languages \cite{schuster2019cross,aldarmaki2019context,wang2020cross,kulshreshtha2020cross,cao2020multilingual}. 

\hspace{3mm} Since a word may appear in different sentences with different contexts, \citet{schuster2019cross} use an average anchor to summarize multiple contexts for a word type and align the anchors of different languages, while other works aim to align each individual context representation based on parallel corpora, including learning alignment on sentence level representations and applying the learned mapping on word level contextual representations \cite{aldarmaki2019context}, using word alignments in a parallel corpora to learn the mapping for word contextual representations \cite{wang2020cross}, and directly minimizing the distance between two contextual representations of an aligned word pair in parallel corpora without mapping \cite{cao2020multilingual}. 

\hspace{3mm} We adopt the average anchor method for the contextual representations \cite{schuster2019cross}, which does not depend on parallel corpora. Let the contextual representation of a source word $x$ in context $c_i$ be denoted as $r_{x, c_i}$. If $x$ appears a total of $p$ times in the source corpus, the average anchor for $x$ across all contexts is: 

\begin{equation}
a_x = \frac{\sum_{i=1}^{p} r_{x,c_i}}{p}   \label{equ:anchor}
\end{equation}

\hspace{3mm} Similar to the mapping for the static word embeddings, we conduct mapping for the average anchors. Let $A_x$ and $A_y$ be the matrices of average anchors in two languages with correspondence to word pairs from a given bilingual dictionary. The mapping functions $V_x$ and $V_y$ are optimized by maximizing $cos(A_x V_x, A_y V_y)$, where $A_x$ and $A_y$ are fixed, $V_x$ and $V_y\in M_{d'}(\mathbb{R})$, and $d'$ is the dimension of the contextual representations.

\end{itemize}

Besides the above methods, there is another direction that extracts word alignments in pseudo parallel corpora for BLI. The pseudo parallel corpora are built by either the unsupervised machine translation \citep{artetxe2019bilingual} or the unsupervised bitext mining \cite{shi2021bilingual}. Both methods need significant computation overload or use monolingual corpora that are magnitudes larger than ours, and are beyond the scope of this paper that focuses on representation based methods.

\section{Proposed Combination Mechanism for BLI}

Since there is no work to combine both the static word embeddings and the contextual representations, we propose a combination mechanism illustrated in Figure \ref{fig:framework}. The mechanism first builds a unified word representation space that unifies the static word embeddings and the contextual representations, then performs similarity interpolation between the unified space and the contextual space.

\subsection{The Unified Word Representations}

As shown in Figure \ref{fig:framework}, the original word embedding space (A) is mapped to (B) through the mapping functions. Since the mapping functions are orthogonal, (A) is just rotated to (B). Notice that the spaces of the two languages are not necessarily isometric everywhere, some words in certain translation pairs are still far away from each other after rotation. To pull words in a translation pair getting closer, we propose a spring network that can pull the mapped embedding points to better positions such that words in a translation pair are nearest neighbors of each other. Since the contextual representations contain rich context information that can be used as the flexible adjustment, the spring network takes the contextual representations as input, and outputs offsets for the word embeddings.

Specifically, in the unified word representations, the mapped word embeddings are pulled to new positions by offsets, which are produced by the spring network with the the contextual representations as input:

\begin{gather}
U_x = E_x^{'} + \gamma_1 \odot F_x(A_x) \notag \\
U_y = E_y^{'} + \gamma_2 \odot F_y(A_y) \label{equ:uni}
\end{gather}

\noindent where $U_x$ and $U_y$ are the unified word representations, $E_x^{'}$ and $E_y^{'}$ are the mapped word embeddings, $F_x$ and $F_y$ are the spring networks, and $\gamma_{1\rm{or}2}$ is the weight vector, which is used to element-wisely multiply each row of the output of the spring network. Take the source side for example, the mapped word embedding matrix $E_x^{'}$ is added with a weighted offset produced by the spring network $F_x$ on the contextual representation (i.e., the average anchor) matrix $A_x$.

~\\
\noindent \textbf{The Spring Network} stacks two feedforward layers with Tanh activations on top of the contextual representation matrices. The first layer transforms the dimension of the contextual representation $d'$ to the dimension of the word embedding $d$. Equations (\ref{equ:layer1}-\ref{equ:layer2}) list the network structure of both sides. 

\begin{align}
A_x^1  &= \phi(\theta_x^0(A_x)), &A_y^1  &=  \phi(\theta_y^0(A_y))  \label{equ:layer1} \\
A_x^2  &= \phi(\theta_x^1(A_x^1)),   &A_y^2 &= \phi(\theta_y^1(A_y^1)) \label{equ:layer2}
\end{align}

\noindent where $\phi$ denotes the Tanh activation, and $\theta$ denotes the feedforward layer. $A_{x/y}^2$ is the output of the spring network, and fulfills  as the offset distance to compensate the deviation of words in each translation pair in the mapped word embedding space. 

Since we use cross-lingual pre-training \citep{lample2019cross} to generate the contextual representations, which are actually BPE's \citep{sennrich2016neural} contextual representations, we have to form the contextual representations in the word level. Suppose a word $x$ has $q$ BPEs, and $x$ appears $p$ times in the monolingual corpus, then the word level contextual representation $a_x = \sum_{i=1}^{p} {( \sum_{j=1}^{q} r_{b_j,c_{i,j} } / {q} ) }  /  {p}$, where $r_{b_j,c_{i,j}}$ denotes the representation of the $j$th BPE with the $i$th context $c_{i,j}$. $a_x$ actually averages $q$ BPEs' representation at first, then averages $p$ contexts. After this cascaded averaging, it constitutes one row of $A_x$.

~\\
\noindent \textbf{Contrastive Training} is used to train the spring networks $F_x$ and $F_y$ with the pre-trained mapped word embeddings and the contextual representations fixed in the unified space. Basically, through the spring adjustment, the training encourages parallel words to get closer, and drives non-parallel words to be farther. It is divided into two scenarios: supervised contrastive training and unsupervised contrastive training.

\begin{itemize}

\item In the supervised contrastive training, given a bilingual dictionary with $I$ translation pairs, the contrastive loss is: 

\begin{align}
\mathcal{L}_{sup} = &- \sum_{i=1}^{I} ( J \times cos(u_x^i, u_y^i)  \notag  \\
 &- \sum_{j=1}^J cos(u_x^i, u_{\bar{y}}^j) )  \label{equ:supervised}
\end{align}

\noindent where $u_x^i$ and $u_y^i$ are the unified representations corresponding to the $i$th entry of the given bilingual dictionary.

\hspace{3mm}  In equation (\ref{equ:supervised}), $(u_x^i, u_y^i)$ is the positive translation pair according to the given dictionary, and the cosine similarity of this pair is maximized during training, while $(u_x^i, u_{\bar{y}}^j)$ is the negative pair where $\bar{y}$ is not aligned to $x$. The cosine similarity of $(u_x^i, u_{\bar{y}}^j)$ is minimized during training.

\hspace{3mm} We select $J$ negative pairs for a source word $x$. In the implementation, we use $J$best outputs of the current model excluding the correct translation as the negative pairs. To keep balance between positive and negative pairs, the positive pair is copied $J$ times to pair with negative pairs. 

\hspace{3mm} During inference, we select $y = \mathop{\arg\max}_y cos(u_x, u_y)$ as the translation of $x$.

\item In the unsupervised contrastive training, no bilingual dictionary is given. The contrastive loss is the same to that of the supervised contrastive training, except that the bilingual dictionary is not given. We initialize the bilingual dictionary using the output of the static word embedding based unsupervised method, and iteratively update it by using the trained model of last iteration to find new translations for given source words and compose a new dictionary, which is used to train the new model. Such process iterates until the dictionary does not change any more. 

\end{itemize}

\subsection{Similarity Interpolation}

The similarity interpolation is for inference. As shown in Figure \ref{fig:framework}, both the unified word representation space and the mapped contextual representation space can output the cosine similarities between words. Given a source word $x$, we interpolate both similarities as below:

\begin{align}
\mathcal{S} = cos(u_x, u_y) + \lambda cos(a_x^{'}, a_y^{'}) \label{equ:inter}
\end{align}

\noindent where $\lambda$ is the weight, $a_{x/y}^{'}$ is the mapped contextual representation, which is pre-trained as introduced in the section of the background of the contextual representation based method. We aim to find $y$ that has the maximal $\mathcal{S}$ as the translation of $x$.

In the supervised setting, $\lambda$ is tuned on the validation set consisting of translation pairs. In the unsupervised setting, $\lambda$ is tuned by an unsupervised procedure: when source-to-target model and target-to-source model have been trained, the word $x$ in the validation set is aligned to $y'$ based on equation (\ref{equ:inter}), then $y'$ is back aligned to $x$ based on the inverse version of equation (\ref{equ:inter}). We select $\lambda$ that has the highest accuracy of this back alignment to $x$. 

\section{Experiments}

We test our combination mechanism in supervised and unsupervised BLI tasks on English-Espanish (EN-ES), English-Arabic (EN-AR), English-Chinese (EN-ZH), English-German (EN-DE), and English-French (EN-FR). 

\begin{table*}[htbp]
\scriptsize
\centering
\begin{tabular}{l|ccccccccccc}
 \bottomrule[1.2pt] 
 & \multicolumn{2}{c}{EN-ES} & \multicolumn{2}{c}{EN-AR} & \multicolumn{2}{c}{EN-ZH} & \multicolumn{2}{c}{EN-DE} & \multicolumn{2}{c}{EN-FR} &  \multirow{2}{*}{avg} \\
 &      $\to$    &  $\gets$           &     $\to$        &       $\gets$       &        $\to$     &  $\gets$            &       $\to$      &         $\gets$     &          $\to$   &   $\gets$   &    \\  \bottomrule[1.2pt] 
 \multicolumn{11}{c}{Supervised BLI} \\ \hline 
 Muse\cite{conneau2017word} &  77.80&     81.40      &    49.47         &    55.07         &  40.87          &    43.07         &    69.80         &    71.07         &    78.67         &   79.33  &  64.65  \\      
 VecMap\cite{artetxe2018aaai} &  77.20&     82.13      &    53.53         &    57.73        &  52.13          &    46.67         &    70.33         &    72.80         &    78.40         &   80.47  & 67.14  \\      
 RCSLS\cite{joulin2018loss} &  79.27&     84.30      &    55.53         &    61.00       &  53.07          &    48.87         &    73.07         &    75.40         &    79.60         &   82.07  & 69.22   \\      
 BLISS\cite{patra2019bilingual} &  79.67&     84.87      &    54.47         &    59.60        &  50.80          &    48.80         &    73.33         &    76.13         &    79.20         &   82.93  & 68.98  \\     \hline 

 Unified$_{\rm {VecMap}}$ & 79.60 & 84.73        &      56.20       &    60.73         & 53.80            &      48.93       & 73.27            &            74.47  &      79.33       & 81.33  & 69.24 \\ 
 Contextual$_{\rm {VecMap}}$ & 44.07 & 50.33 & 5.07 & 7.73  & 21.93 & 9.67 &44.60  &47.87  & 57.47 & 65.20  & 35.39 \\
 Interpolation$_{\rm {VecMap}}$  & 80.47 & 85.70 & 57.13 & 61.47 & 56.27 & 50.60 & 74.13 & 77.13 & 80.80 & 83.40 & 70.71 \\  \hline
 
 Unified$_{\rm {RCSLS}}$ & 80.13 & 86.60 & 55.87 & 62.47  & 56.67 & 51.13 & 74.26 & 77.93 & 81.20  &  83.87 &  71.01  \\
 Contextual$_{\rm {RCSLS}}$ & 46.27 & 51.40 & 3.67 & 7.13  & 19.47 &  7.93 &45.67  & 47.80 &  58.00 &65.20 & 35.25   \\
 Interpolation$_{\rm {RCSLS}}$ & \textbf{80.67} & \textbf{87.67} & \textbf{59.40} & \textbf{62.73}  & \textbf{59.40} & \textbf{52.27} & \textbf{74.87} & \textbf{79.67} & \textbf{81.80} & \textbf{85.40}  & \textbf{72.39}  \\ 
 
\bottomrule[1.2pt]
\multicolumn{11}{c}{Unsupervised BLI} \\ \hline 
 Muse\cite{conneau2017word} & 77.06 &81.53  &48.00  & 55.47  & 24.26 & 43.00 & 70.13 & 71.20 & 78.73 &    78.40 & 62.77  \\
 VecMap\cite{artetxe2018acl} & 77.60 &81.67  &50.87  & 56.73  & 34.33 & 44.00 & 70.00 & 71.80 & 78.73 &    80.27 & 64.60 \\
 Ad.\cite{mohiuddin19naacl} &77.93&82.20&50.07&57.33& 34.67&43.67&69.13&72.47&78.46&80.13&64.61 \\  \hline
 Unified & 79.47  &     82.60        &    52.47         &    57.87         & 35.93            &      46.07       & 71.07            &            74.00  &      80.27       & 80.87 & 66.06  \\ 
 Contextual & 46.33& 55.93 &3.87  &7.53  & 17.40  & 4.87   & 43.53 &  44.33&56.20  &  63.93  & 34.39 \\
 Interpolation & \textbf{79.93} & \textbf{85.33} & \textbf{52.73} & \textbf{58.47} & \textbf{37.07} & \textbf{46.27} & \textbf{72.53} & \textbf{78.73} & \textbf{81.80} & \textbf{84.13}  &  \textbf{67.70} \\ 
  \toprule[1.2pt]
\end{tabular} 
\caption{
P@1 on all language pairs. ``Unified'' denotes our unified word representation based method, which computes $cos(u_x, u_y)$,  ``Contextual'' denotes the contextual representation based method, which computes $cos(a_x^{'}, a_y^{'})$, ``Interpolation'' denotes our similarity interpolation, which computes $cos(u_x, u_y) + \lambda cos(a_x^{'}, a_y^{'})$. The subscript ``${\rm {VecMap}}$'' denotes that our method is based on the work of \citet{artetxe2018aaai}, the subscript ``$\rm RCSLS$'' denotes that our method is based on the RCSLS criterion in training \cite{joulin2018loss}. In unsupervised BLI, there is no subscript in our method, which means using the default ``$\rm VecMap$'' \citep{artetxe2018acl}.
}
\label{tbl:mainResult}
\end{table*}

\subsection{Data}

We need monolingual corpora to compute the contextual representations. Unfortunately, most existing BLI datasets distribute pre-trained word embeddings alone, but not the monolingual corpora used to train them. For that reason, we use WikiExtractor\footnote{https://github.com/attardi/wikiextractor} to extract plain text from Wikipedia dumps, and preprocess the resulting corpora using standard Moses \citep{koehn2007moses} tools by applying sentence splitting, punctuation normalization, tokenization, and lowercasing. On these corpora, we use the cross-lingual pre-training system XLM \cite{lample2019cross}\footnote{https://github.com/facebookresearch/xlm. We use the MLM model of 15 languages with tokenize + lowercase + no accent + BPE.} to compute the contextual representations.

Meanwhile, we also use these corpora to train the static word embeddings by using fastText\footnote{https://github.com/facebookresearch/fastText/} to ensure that both the contextual representations and the static word embeddings come from the same data. We use the bilingual dictionaries released by Muse project\footnote{https://github.com/facebookresearch/MUSE} in our experiments. Note that some words in these dictionaries do not necessarily appear in our monolingual corpora, we have to recompose the training, validation, and test sets such that all words in these sets are included in our monolingual corpora. In the end, we have 5000 entries with unique source words in the training set, and 1500 entries with unique source words in both the validation set and the test set for all language pairs. 

\subsection{Baseline Systems}

Baseline systems are divided into two tasks as below. We run the released code of each baseline system in our experiments. 

\noindent \textbf{Supervised BLI task}, which is allowed to use bilingual dictionaries for training and validation. The baseline systems are:

\begin{itemize}

\item \textbf{Muse}: Supervised Muse is set as the baseline in \citet{conneau2017word}. It uses iterative Procrustes alignment for supervised BLI.

\item \textbf{VecMap}\footnote{https://github.com/artetxem/vecmap}: \citet{artetxe2018aaai} use a multi-step framework consisting of several steps: whitening, orthogonal mapping, re-weighting, de-whitening, and dimensionality reduction.

\item \textbf{RCSLS}\footnote{\url{https://github.com/facebookresearch/fastText/tree/master/alignment} }: In addition to use CSLS during inference, \citet{joulin2018loss} minimize a convex relaxation of CSLS loss during training, and improve the supervised BLI performance.


\item \textbf{BLISS}\footnote{https://github.com/joelmoniz/BLISS}: \citet{patra2019bilingual} use a semi-supervised method that leverages both the bilingual dictionary and a larger set of unaligned word embeddings.

\end{itemize}

\noindent \textbf{Unsupervised BLI task}, which is not allowed to use any parallel resources for training and validation. The baseline systems are: 

\begin{itemize}

\item \textbf{Muse}: Unsupervised Muse \citep{conneau2017word} uses adversarial training and iterative Procrustes refinement. 

\item \textbf{VecMap}: \citet{artetxe2018acl} use careful initialization, robust self-learning procedure, and symmetric re-weighting to improve the unsupervised mapping result. 

\item \textbf{Ad.}\footnote{https://github.com/taasnim/unsup-word-translation/}: \citet{mohiuddin19naacl} include regularization terms for adversarial auto-encoder for the unsupervised BLI.

\end{itemize}

\subsection{Experimental Settings}

We use fastText to train the word embeddings for BLI. The dimension of the word embeddings is 300. The contextual representations are extracted from XLM, and the dimension of the contextual representations is 1024. For each word type, we randomly select ten sentences containing the word from the monolingual corpora to do the averaging to get the contextual representation. The influence of the number of selected sentences for each word type is reported in section \ref{sec:random}. Regarding the spring network, we use ten negative pairs for each source word in the supervised contrastive training, and use one negative pair for each source word in the unsupervised contrastive training. 

All inferences in our experiments, including all baseline systems, use CSLS which is introduced in \citet{conneau2017word}. The results are evaluated by Precision@1 (P@1). 

\begin{table*}[htbp]
\small
\centering
\begin{tabular}{l|ccccccccccc}
 \bottomrule[1.2pt] 
 & \multicolumn{2}{c}{EN-ES} & \multicolumn{2}{c}{EN-AR} & \multicolumn{2}{c}{EN-ZH} & \multicolumn{2}{c}{EN-DE} & \multicolumn{2}{c}{EN-FR} &  \multirow{2}{*}{avg} \\
 &      $\to$    &  $\gets$           &     $\to$        &       $\gets$       &        $\to$     &  $\gets$            &       $\to$      &         $\gets$     &          $\to$   &   $\gets$   &    \\   \bottomrule[1.2pt] 
 \multicolumn{11}{c}{Supervised BLI} \\ \hline     
 XLM-Unified$_{\rm {RCSLS}}$  & 80.13 & 86.60 & 55.87 & 62.47  & 56.67 & 51.13 & 74.26 & 77.93 & 81.20  &  83.87 &  71.01   \\
 XLM-Contextual$_{\rm {RCSLS}}$ & 46.27 & 51.40 & 3.67 & 7.13  & 19.47 &  7.93 &45.67  & 47.80 &  58.00 &65.20 & 32.05   \\   
 XLM-Interpolated$_{\rm {RCSLS}}$ & {80.67} & \textbf{87.67} & \textbf{59.40} & {62.73}  & \textbf{59.40} & \textbf{52.27} & {74.87} & {79.67} & {81.80} & {85.40}  & \textbf{72.39}   \\    \hline
 
 mBART-Unified$_{\rm {RCSLS}}$ & 80.07  &    85.33       &     56.13        &    61.93    &    55.53   &   51.20    &     74.27     &   77.53     &  80.20      &   83.40   &  70.56     \\  
 mBART-Contextual$_{\rm {RCSLS}}$ & 46.33 & 49.70 & 3.53 & 8.24  & 18.80 &  8.03 &45.33  & 48.33 &  60.10 &66.53  & 32.26  \\   
 mBART-Interpolated$_{\rm {RCSLS}}$ &  \textbf{81.27}  &     87.40   &    57.00  &  \textbf{63.00}   &     58.27  &  52.07 &  \textbf{76.07}    &  \textbf{80.60}   &  \textbf{82.33}   &   \textbf{85.93}   &   \textbf{72.39}   \\	
 
 \bottomrule[1.2pt]
\multicolumn{11}{c}{Unsupervised BLI} \\ \hline 
 XLM-Unified & 79.47  &     82.60        &    52.47         &    57.87         & 35.93            &      46.07       & 71.07            &            74.00  &      80.27       & 80.87 & 66.06      \\   
 XLM-Contextual & 46.33& 55.93 &3.87  &7.53  & 17.40  & 4.87   & 43.53 &  44.33&56.20  &  63.93  & 31.26 \\
 XLM-Interpolated & \textbf{79.93} & \textbf{85.33} & \textbf{52.73} & {58.47} & \textbf{37.07} & \textbf{46.27} & \textbf{72.53} & \textbf{78.73} & \textbf{81.80} & {84.13}  &  \textbf{67.70}  \\     \hline
 
 mBART-Unified & 79.07  &     82.60      &    51.60      &     58.13       &    35.80        &      44.67       &     70.47  &  74.00     &    79.93     &    81.27  &  65.75    \\  
mBART-Contextual & 47.06& 53.37 &5.40  &8.13  & 16.76  & 4.77   & 41.23 &  42.53&53.53  &  62.87  & 30.51 \\    
 mBART-Interpolated &  \textbf{79.93} &     84.90      &    \textbf{52.73}     &    \textbf{59.67}        &    36.60        &      45.27       &     71.87        &      76.33       &      81.40       &   \textbf{84.53}   &   67.32   \\  
  \toprule[1.2pt]
\end{tabular} 
\caption{
Result comparison between using XLM and using mBART for the contextual representations.
}
\label{tbl:mbart-xlm}
\end{table*}

\subsection{Main Results}

Table \ref{tbl:mainResult} summarizes the main results of the supervised and the unsupervised BLI tasks on all test sets. In both tasks, our proposed methods achieve significant improvements, with average 3.2 points higher than the strongest baseline RCSLS in the supervised task, and with average 3.1 points higher than the strong baselines VecMap and Ad. in the unsupervised task.

\begin{figure}[htbp]
\flushleft
\centerline {\includegraphics[height=6cm,width=8cm]{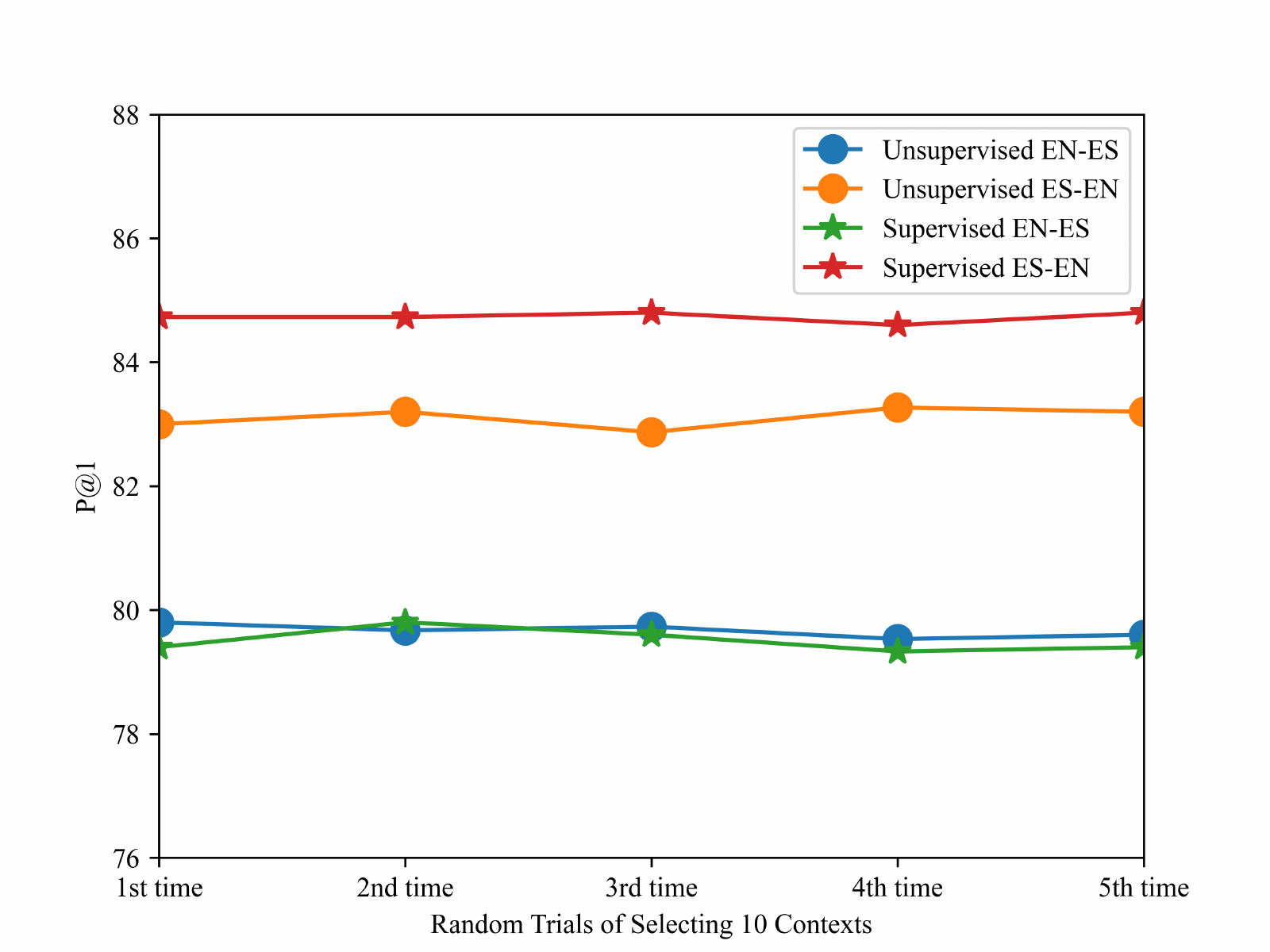} }
\caption{
Performances of 5 random trials of selecting 10 sentences to gather contexts for each word type.
}
\label{fig:trials}
\end{figure}

In the supervised task, we have two independent bases to build our proposed methods. One is VecMap \citep{artetxe2018aaai}, the other is RCSLS \citep{joulin2018loss}. They are the preprocessing steps  to align the static word embeddings, and align the contextual representations in the two languages. Our methods build upon these alignments, and further train the spring networks and the unified word representations for the combination. The performances of our methods with these two bases are reported in Table \ref{tbl:mainResult} with the corresponding subscripts. 

Table \ref{tbl:mainResult} shows that if we use VecMap as the basis of our method, we can improve 3.6 points over the corresponding VecMap baseline. If we use RCSLS as the basis, we can improve 3.2 points over the corresponding RCSLS baseline. In our methods, ``Unified'' can achieve around 2 points improvement over the corresponding baselines. Although ``Contextual'' obtains inferior performances, it is complementary to ``Unified''. When ``Contextual'' is combined with ``Unified'' through the interpolation, the performance is further improved, achieving the best performance among all systems. It shows that our combination mechanism is effective to utilize the merits of both the static word embeddings and the contextual representations. 

In the unsupervised task, we achieve the significant improvements over the baselines. ``Unified'' is 1.5 points better than VecMap baseline. ``Contextual'' is inferior to other methods, but it can provide useful complements to ``Unified'', resulting in the final 3.1 points improvement through interpolation. 

In summary, our combination mechanism consistently improves the performances for both distant language pairs, such as EN-AR and EN-ZH, and closely-related European language pairs.

\subsection{Analyses}

\subsubsection{XLM v.s. mBART}

Our results in Table \ref{tbl:mainResult} are based on using XLM for obtaining the contextual representations. In this section, we also use mBART  \citep{Liu2020Multilingual} to compare with XLM. Table \ref{tbl:mbart-xlm} shows the comparison result. XLM pre-trains the Transformer encoder through the masking mechanism, while mBART pre-trains the full Transformer encoder-decoder through multilingual denoising. Regarding BLI task, we obtain the contextual representations from the encoder. Table \ref{tbl:mbart-xlm} shows that XLM and mBART get similar BLI performances since only encoder is used. In some directions, mBART performs slightly better than XLM, while in other directions, XLM is slightly better. According to the average performance, XLM ties with mBART in the supervised task, and is slightly better in the unsupervised task.

\subsubsection{The Randomness of Contexts} \label{sec:random}

\begin{figure}[htbp]
\flushleft
\centerline {\includegraphics[height=6cm,width=8cm]{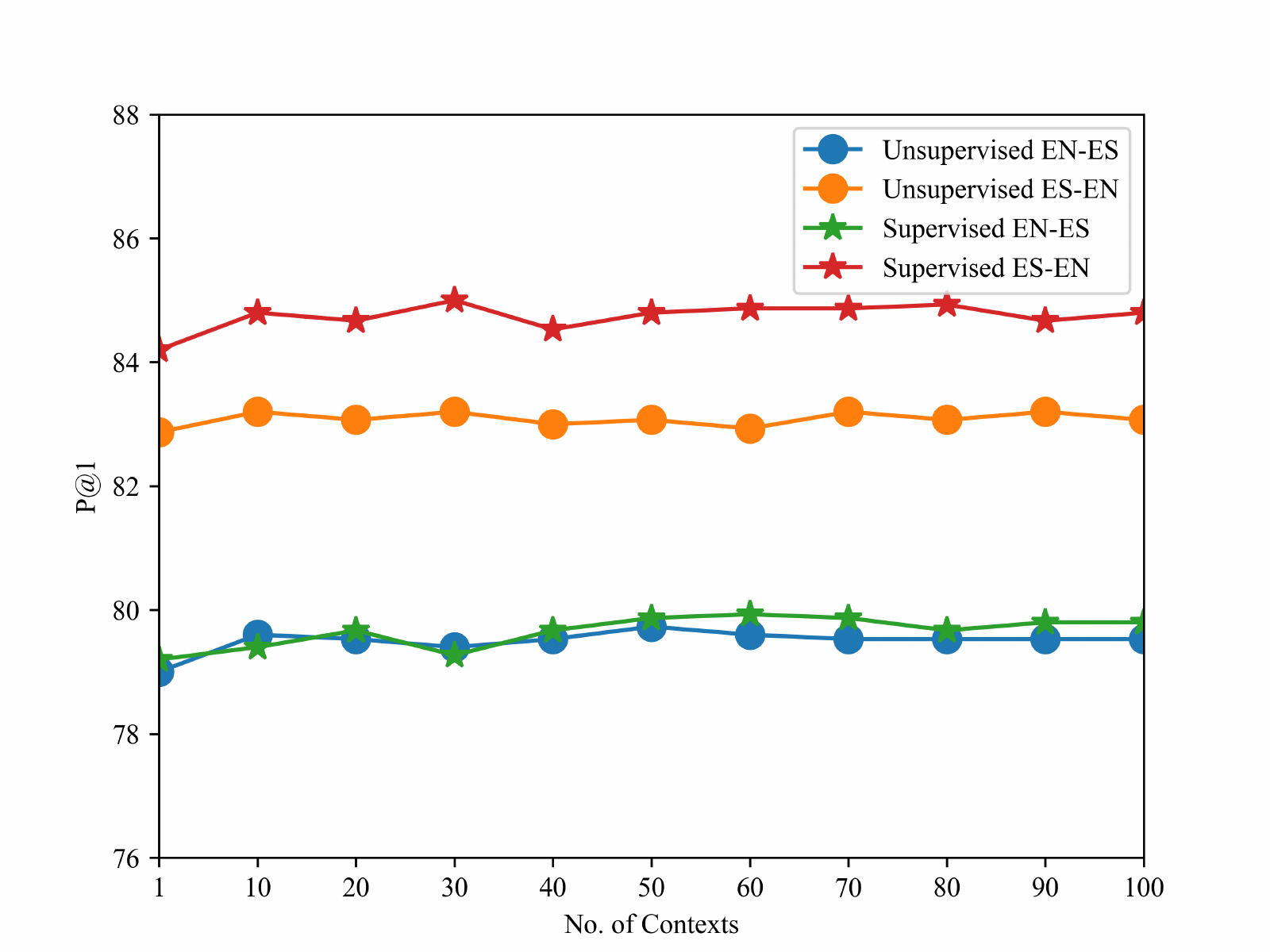} }
\caption{
Performances of randomly selecting 1-100 sentences to gather contexts for each word type.
}
\label{fig:context}
\end{figure}

\begin{table}[htbp]
\small
\centering
\begin{tabular}{l|cccc}
 \bottomrule[1.2pt] 
 & \multicolumn{2}{c}{EN-DE} & \multicolumn{2}{c}{EN-FR}  \\
 &      $\to$    &  $\gets$           &     $\to$        &       $\gets$       \\   \bottomrule[1.2pt] 
 \multicolumn{4}{c}{Supervised BLI} \\ \hline    
 RCSLS &  61.82  &     60.92      &      59.58      &     60.92       \\   \hline
 Unified &  62.93   &     62.35      &       61.87     &       62.40        \\   
 Interpolation &   \textbf{67.09}   &     \textbf{68.00}      &     \textbf{67.05}       &    \textbf{67.12}    \\   
 
 \bottomrule[1.2pt]
 \multicolumn{4}{c}{Unsupervised BLI} \\ \hline 
 VecMap & 55.29  &      56.96     &     57.18       &      59.33          \\   \hline
 Unified &      57.20     &       58.70     &        58.40   &   61.16        \\   
 Interpolation  &   \textbf{61.00}    &    \textbf{61.52}       &      \textbf{62.70}      &     \textbf{63.40}        \\   
 
  \toprule[1.2pt]
\end{tabular} 
\caption{
P@1 of using WaCKy corpora.
}
\label{tbl:wacky}
\end{table}

The contextual representations are derived randomly from sentences of the monolingual corpora. We study if this random derivation affects the performances. Firstly, we run 5 times of randomly selecting 10 sentences to gather contexts for each word type in the ``Unified'' setting. Figure \ref{fig:trials} shows that the performance is stable in the 5 trials. Secondly, we try randomly selecting 1-100 sentences to gather contexts for each word type. Figure \ref{fig:context} shows that selecting 1 sentence will drag the performance down to baseline, which indicates that 1 sentence is too random to gather enough information for BLI.

We only list the studies on EN-ES due to space limit. Studies on other language pairs can be found in the appendix. 

\subsubsection{The Influence of Selecting Encoder Layer} \label{sec:layers}

\begin{figure}[htbp]
\flushleft
\centerline {\includegraphics[height=6cm,width=8cm]{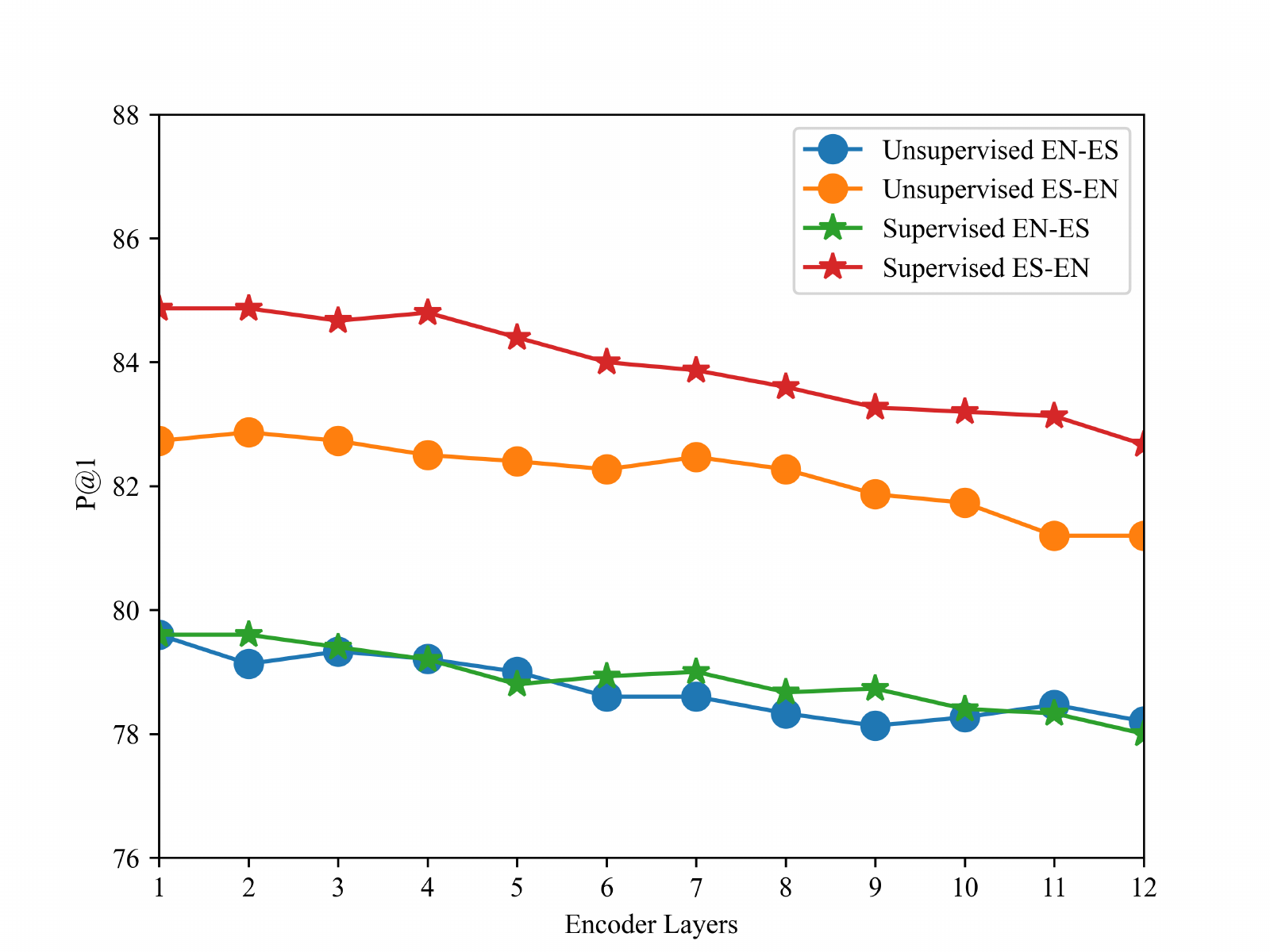} }
\caption{
Performances of selecting different layer of XLM encoder to derive the contextual representations.
}
\label{fig:layers}
\end{figure}

In the above experiments, we derive the contextual representations from the first layer of the encoder of XML/mBART. In this section, we show how different will be when we change the layer in the ``Unified'' setting. Figure \ref{fig:layers} shows that as layer go higher, the performance drops. Please refer to the appendix for performances of other language pairs.

\subsubsection{Results of using WaCKy Corpora}

WaCKy corpora is introduced in \citet{dinu2014improving} for BLI, but only word embeddings trained on WaCKy corpora are provided in their work. To obtain the contextual representations, we find WaCKy corpora from BUCC\footnote{https://comparable.limsi.fr/bucc2020/bucc2020-task.html}, and use the corresponding dictionaries with the same training, validation, and test split. We use mBART instead of XLM for computing the contextual representations in this task.



Table \ref{tbl:wacky} shows that our combination mechanism is robust on this dataset. Both ``Unified'' and ``Interpolation'' perform better than the baselines. ``Interpolation'' achieves significant improvements in the supervised setting.

\section{Discussion}

The static word embeddings in our paper are trained by skip-gram or CBOW, while the word embeddings from XLM/mBART are trained by the pre-training objectives. Different training objectives result in quite different word embeddings. Actually, they show remarkably different behavior for BLI. By using VecMap, the word embeddings from XLM/mBART perform averagely around 30 points lower than the static word embeddings used in our paper. We also test fastText with 1024 dimension and word2vec with 300 dimension for fair comparison. They all perform remarkably better than the word embeddings from XLM/mBART. We plug in the word embeddings from XLM/mBART in place of the fastText static embeddings in our combination approach, and obtain much worse performance. This indicates that the static word embeddings trained by skip-gram or CBOW are more suitable for BLI and our combination approach.

In addition, regarding the asymmetry in Figure \ref{fig:framework}, we used to try the spring function that takes the static word embeddings as input, but got much worse results. This indicates that the spring function conditioned on the static space may be not helpful for BLI. Such observation may also explain that symmetrizing Figure \ref{fig:framework} by yielding two unified spaces to combine performs slightly worse than the asymmetry version of Figure 1 of this paper. This is because that the spring function conditioned on the static space is introduced to maintain the symmetry, while this introduced spring function is not helpful for the combination.

\section{Conclusion}

Most BLI systems use either the static word embeddings or the contextual representations, but there is no works to combine both. In this paper, we propose a combination mechanism, which consists of the unified word representations and the similarity interpolation. The unified word representations use a spring network to pull the static word embeddings with offsets produced by the contextual representations, and compose a unified space such that parallel words are nearest neighbors to each other. The similarity interpolation is applied afterward to interpolate the similarities in the unified space and the contextual representation space. BLI experiments on multiple language pairs show that our combination mechanism can utilize the merits of both the static word embeddings and the contextual representations, achieving significant improvements over robust baseline systems in both the supervised and the unsupervised BLI tasks.

\section*{Acknowledgments}

The authors would like to thank the anonymous reviewers for the helpful comments. This work was supported by National Natural Science Foundation of China (Grant No. 62036004, 61673289)

\bibliographystyle{acl_natbib}
\bibliography{acl2021}

\begin{thebibliography}{38}
\expandafter\ifx\csname natexlab\endcsname\relax\def\natexlab#1{#1}\fi

\bibitem[{Aldarmaki and Diab(2019)}]{aldarmaki2019context}
Hanan Aldarmaki and Mona Diab. 2019.
\newblock Context-aware cross-lingual mapping.
\newblock In \emph{Proceedings of the 2019 Conference of the North {A}merican
  Chapter of the Association for Computational Linguistics: Human Language
  Technologies, Volume 1 (Long and Short Papers)}, pages 3906--3911.

\bibitem[{Artetxe et~al.(2016)Artetxe, Labaka, and
  Agirre}]{artetxe2016learning}
Mikel Artetxe, Gorka Labaka, and Eneko Agirre. 2016.
\newblock Learning principled bilingual mappings of word embeddings while
  preserving monolingual invariance.
\newblock In \emph{Proceedings of the 2016 Conference on Empirical Methods in
  Natural Language Processing}, pages 2289--2294.

\bibitem[{Artetxe et~al.(2017)Artetxe, Labaka, and
  Agirre}]{artetxe2017learning}
Mikel Artetxe, Gorka Labaka, and Eneko Agirre. 2017.
\newblock Learning bilingual word embeddings with (almost) no bilingual data.
\newblock In \emph{Proceedings of the 55th Annual Meeting of the Association
  for Computational Linguistics (Volume 1: Long Papers)}, pages 451--462.

\bibitem[{Artetxe et~al.(2018{\natexlab{a}})Artetxe, Labaka, and
  Agirre}]{artetxe2018aaai}
Mikel Artetxe, Gorka Labaka, and Eneko Agirre. 2018{\natexlab{a}}.
\newblock Generalizing and improving bilingual word embedding mappings with a
  multi-step framework of linear transformations.
\newblock In \emph{Proceedings of the Thirty-Second AAAI Conference on
  Artificial Intelligence}, pages 5012--5019.

\bibitem[{Artetxe et~al.(2018{\natexlab{b}})Artetxe, Labaka, and
  Agirre}]{artetxe2018acl}
Mikel Artetxe, Gorka Labaka, and Eneko Agirre. 2018{\natexlab{b}}.
\newblock A robust self-learning method for fully unsupervised cross-lingual
  mappings of word embeddings.
\newblock In \emph{Proceedings of the 56th Annual Meeting of the Association
  for Computational Linguistics (Volume 1: Long Papers)}, pages 789--798.

\bibitem[{Artetxe et~al.(2019)Artetxe, Labaka, and
  Agirre}]{artetxe2019bilingual}
Mikel Artetxe, Gorka Labaka, and Eneko Agirre. 2019.
\newblock Bilingual lexicon induction through unsupervised machine translation.
\newblock In \emph{Proceedings of the 57th Annual Meeting of the Association
  for Computational Linguistics}, pages 5002--5007.

\bibitem[{Cao et~al.(2020)Cao, Kitaev, and Klein}]{cao2020multilingual}
Steven Cao, Nikita Kitaev, and Dan Klein. 2020.
\newblock Multilingual alignment of contextual word representations.
\newblock In \emph{Proceedings of International Conference on Learning
  Representations}.

\bibitem[{Conneau et~al.(2017)Conneau, Lample, Ranzato, Denoyer, and
  J{\'e}gou}]{conneau2017word}
Alexis Conneau, Guillaume Lample, Marc'Aurelio Ranzato, Ludovic Denoyer, and
  Herv{\'e} J{\'e}gou. 2017.
\newblock Word translation without parallel data.
\newblock \emph{arXiv preprint arXiv:1710.04087}.

\bibitem[{Devlin et~al.(2019)Devlin, Chang, Lee, and
  Toutanova}]{devlin-etal-2019-bert}
Jacob Devlin, Ming-Wei Chang, Kenton Lee, and Kristina Toutanova. 2019.
\newblock {BERT}: Pre-training of deep bidirectional transformers for language
  understanding.
\newblock In \emph{Proceedings of the 2019 Conference of the North {A}merican
  Chapter of the Association for Computational Linguistics: Human Language
  Technologies, Volume 1 (Long and Short Papers)}, pages 4171--4186.

\bibitem[{Dinu et~al.(2014)Dinu, Lazaridou, and Baroni}]{dinu2014improving}
Georgiana Dinu, Angeliki Lazaridou, and Marco Baroni. 2014.
\newblock Improving zero-shot learning by mitigating the hubness problem.
\newblock \emph{arXiv preprint arXiv:1412.6568}.

\bibitem[{Fung and Yee(1998)}]{fung1998ir}
Pascale Fung and Lo~Yuen Yee. 1998.
\newblock An ir approach for translating new words from nonparallel, comparable
  texts.
\newblock In \emph{36th Annual Meeting of the Association for Computational
  Linguistics and 17th International Conference on Computational Linguistics,
  Volume 1}, pages 414--420.

\bibitem[{Haghighi et~al.(2008)Haghighi, Liang, Berg-Kirkpatrick, and
  Klein}]{haghighi2008learning}
Aria Haghighi, Percy Liang, Taylor Berg-Kirkpatrick, and Dan Klein. 2008.
\newblock Learning bilingual lexicons from monolingual corpora.
\newblock In \emph{Proceedings of the Annual Meeting of the Association for
  Computational Linguistics 2008: HLT}, pages 771--779.

\bibitem[{Irvine and Callison-Burch(2013)}]{irvine2013combining}
Ann Irvine and Chris Callison-Burch. 2013.
\newblock Combining bilingual and comparable corpora for low resource machine
  translation.
\newblock In \emph{Proceedings of the eighth workshop on statistical machine
  translation}, pages 262--270.

\bibitem[{Joulin et~al.(2018)Joulin, Bojanowski, Mikolov, J{\'e}gou, and
  Grave}]{joulin2018loss}
Armand Joulin, Piotr Bojanowski, Tomas Mikolov, Herv{\'e} J{\'e}gou, and
  Edouard Grave. 2018.
\newblock Loss in translation: Learning bilingual word mapping with a retrieval
  criterion.
\newblock In \emph{Proceedings of the 2018 Conference on Empirical Methods in
  Natural Language Processing}, pages 2979--2984.

\bibitem[{Koehn et~al.(2007)Koehn, Hoang, Birch, Callison-Burch, Federico,
  Bertoldi, Cowan, Shen, Moran, Zens, Dyer, Bojar, Constantin, and
  Herbst}]{koehn2007moses}
Philipp Koehn, Hieu Hoang, Alexandra Birch, Chris Callison-Burch, Marcello
  Federico, Nicola Bertoldi, Brooke Cowan, Wade Shen, Christine Moran, Richard
  Zens, Chris Dyer, Ond{\v{r}}ej Bojar, Alexandra Constantin, and Evan Herbst.
  2007.
\newblock {M}oses: Open source toolkit for statistical machine translation.
\newblock In \emph{Proceedings of the 45th Annual Meeting of the Association
  for Computational Linguistics Companion Volume Proceedings of the Demo and
  Poster Sessions}, pages 177--180.

\bibitem[{Kulshreshtha et~al.(2020)Kulshreshtha, Redondo-Garc{\'\i}a, and
  Chang}]{kulshreshtha2020cross}
Saurabh Kulshreshtha, Jos{\'e}~Luis Redondo-Garc{\'\i}a, and Ching-Yun Chang.
  2020.
\newblock Cross-lingual alignment methods for multilingual bert: A comparative
  study.
\newblock \emph{arXiv preprint arXiv:2009.14304}.

\bibitem[{Lample and Conneau(2019)}]{lample2019cross}
Guillaume Lample and Alexis Conneau. 2019.
\newblock Cross-lingual language model pretraining.
\newblock In \emph{Proceedings of the Advances in Neural Information Processing
  Systems (NeurIPS)}, pages 7059--7069.

\bibitem[{Liu et~al.(2020)Liu, Gu, Goyal, Li, Edunov, Ghazvininejad, Lewis, and
  Zettlemoyer}]{Liu2020Multilingual}
Yinhan Liu, Jiatao Gu, Naman Goyal, Xian Li, Sergey Edunov, Marjan
  Ghazvininejad, Mike Lewis, and Luke Zettlemoyer. 2020.
\newblock Multilingual denoising pre-training for neural machine translation.
\newblock \emph{arXiv preprint arXiv:2001.08210}.

\bibitem[{Mikolov et~al.(2013)Mikolov, Le, and
  Sutskever}]{mikolov2013exploiting}
Tomas Mikolov, Quoc~V Le, and Ilya Sutskever. 2013.
\newblock Exploiting similarities among languages for machine translation.
\newblock \emph{arXiv preprint arXiv:1309.4168}.

\bibitem[{Mohiuddin et~al.(2020)Mohiuddin, Bari, and Joty}]{mohiuddin2020lnmap}
Tasnim Mohiuddin, M~Saiful Bari, and Shafiq Joty. 2020.
\newblock Lnmap departures from isomorphic assumption in bilingual lexicon
  induction through non-linear mapping in latent space.
\newblock In \emph{Proceedings of the 2020 Conference on Empirical Methods in
  Natural Language Processing}, pages 2712--2723.

\bibitem[{Mohiuddin and Joty(2019)}]{mohiuddin19naacl}
Tasnim Mohiuddin and Shafiq Joty. 2019.
\newblock {Revisiting Adversarial Autoencoder for Unsupervised Word Translation
  with Cycle Consistency and Improved Training}.
\newblock In \emph{Proceedings of the 2019 Conference of the North American
  Chapter of the Association for Computational Linguistics: Human Language
  Technologies}, pages 3857--3867.

\bibitem[{Patra et~al.(2019)Patra, Moniz, Garg, Gormley, and
  Neubig}]{patra2019bilingual}
Barun Patra, Joel Ruben~Antony Moniz, Sarthak Garg, Matthew~R Gormley, and
  Graham Neubig. 2019.
\newblock Bilingual lexicon induction with semi-supervision in non-isometric
  embedding spaces.
\newblock In \emph{Proceedings of the 57th Annual Meeting of the Association
  for Computational Linguistics}, pages 184--193.

\bibitem[{Qi et~al.(2018)Qi, Sachan, Felix, Padmanabhan, and
  Neubig}]{qi2018and}
Ye~Qi, Devendra~Singh Sachan, Matthieu Felix, Sarguna~Janani Padmanabhan, and
  Graham Neubig. 2018.
\newblock When and why are pre-trained word embeddings useful for neural
  machine translation?
\newblock \emph{arXiv preprint arXiv:1804.06323}.

\bibitem[{Rapp(1999)}]{rapp1999automatic}
Reinhard Rapp. 1999.
\newblock Automatic identification of word translations from unrelated english
  and german corpora.
\newblock In \emph{Proceedings of the 37th annual meeting of the Association
  for Computational Linguistics}, pages 519--526.

\bibitem[{Ravi and Knight(2011)}]{ravi2011deciphering}
Sujith Ravi and Kevin Knight. 2011.
\newblock Deciphering foreign language.
\newblock In \emph{Proceedings of the 49th Annual Meeting of the Association
  for Computational Linguistics: Human Language Technologies}, pages 12--21.

\bibitem[{Ruder et~al.(2019)Ruder, Vuli{\'c}, and S{\o}gaard}]{ruder2019survey}
Sebastian Ruder, Ivan Vuli{\'c}, and Anders S{\o}gaard. 2019.
\newblock A survey of cross-lingual word embedding models.
\newblock \emph{Journal of Artificial Intelligence Research}, 65:569--631.

\bibitem[{Schuster et~al.(2019)Schuster, Ram, Barzilay, and
  Globerson}]{schuster2019cross}
Tal Schuster, Ori Ram, Regina Barzilay, and Amir Globerson. 2019.
\newblock Cross-lingual alignment of contextual word embeddings, with
  applications to zero-shot dependency parsing.
\newblock In \emph{Proceedings of the 2019 Conference of the North {A}merican
  Chapter of the Association for Computational Linguistics: Human Language
  Technologies, Volume 1 (Long and Short Papers)}, pages 1599--1613.

\bibitem[{Sennrich et~al.(2016)Sennrich, Haddow, and
  Birch}]{sennrich2016neural}
Rico Sennrich, Barry Haddow, and Alexandra Birch. 2016.
\newblock Neural machine translation of rare words with subword units.
\newblock In \emph{Proceedings of the 54th Annual Meeting of the Association
  for Computational Linguistics (Volume 1: Long Papers)}, pages 1715--1725.

\bibitem[{Shi et~al.(2021)Shi, Zettlemoyer, and Sida}]{shi2021bilingual}
Haoyue Shi, Luke Zettlemoyer, and Ikbal Sida. 2021.
\newblock Bilingual lexicon induction via unsupervised bitext construction and
  word alignment.
\newblock \emph{arXiv preprint arXiv:2101.00148}.

\bibitem[{Smith et~al.(2017)Smith, Turban, Hamblin, and
  Hammerla}]{smith2017offline}
Samuel~L Smith, David~HP Turban, Steven Hamblin, and Nils~Y Hammerla. 2017.
\newblock Offline bilingual word vectors, orthogonal transformations and the
  inverted softmax.
\newblock \emph{arXiv preprint arXiv:1702.03859}.

\bibitem[{S{\o}gaard et~al.(2018)S{\o}gaard, Ruder, and
  Vuli{\'c}}]{sogaard-etal-2018-limitations}
Anders S{\o}gaard, Sebastian Ruder, and Ivan Vuli{\'c}. 2018.
\newblock On the limitations of unsupervised bilingual dictionary induction.
\newblock In \emph{Proceedings of the 56th Annual Meeting of the Association
  for Computational Linguistics (Volume 1: Long Papers)}, pages 778--788.

\bibitem[{Vuli{\'c} and Moens(2013)}]{vulic2013study}
Ivan Vuli{\'c} and Marie~Francine Moens. 2013.
\newblock A study on bootstrapping bilingual vector spaces from non-parallel
  data (and nothing else).
\newblock In \emph{Proceedings of the 2013 Conference on Empirical Methods in
  Natural Language Processing}, pages 1613--1624.

\bibitem[{Wang et~al.(2020)Wang, Xie, Xu, Yang, Neubig, and
  Carbonell}]{wang2020cross}
Zirui Wang, Jiateng Xie, Ruochen Xu, Yiming Yang, Graham Neubig, and Jaime
  Carbonell. 2020.
\newblock Cross-lingual alignment vs joint training: A comparative study and a
  simple unified framework.
\newblock In \emph{Proceedings of International Conference on Learning
  Representations}.

\bibitem[{Xiao and Guo(2014)}]{xiao2014distributed}
Min Xiao and Yuhong Guo. 2014.
\newblock Distributed word representation learning for cross-lingual dependency
  parsing.
\newblock In \emph{Proceedings of the Eighteenth Conference on Computational
  Natural Language Learning}, pages 119--129.

\bibitem[{Xing et~al.(2015)Xing, Wang, Liu, and Lin}]{xing2015normalized}
Chao Xing, Dong Wang, Chao Liu, and Yiye Lin. 2015.
\newblock Normalized word embedding and orthogonal transform for bilingual word
  translation.
\newblock In \emph{Proceedings of the 2015 Conference of the North American
  Chapter of the Association for Computational Linguistics: Human Language
  Technologies}, pages 1006--1011.

\bibitem[{Zhang et~al.(2017{\natexlab{a}})Zhang, Liu, Luan, and
  Sun}]{zhang2017adversarial}
Meng Zhang, Yang Liu, Huanbo Luan, and Maosong Sun. 2017{\natexlab{a}}.
\newblock Adversarial training for unsupervised bilingual lexicon induction.
\newblock In \emph{Proceedings of the 55th Annual Meeting of the Association
  for Computational Linguistics (Volume 1: Long Papers)}, pages 1959--1970.

\bibitem[{Zhang et~al.(2017{\natexlab{b}})Zhang, Liu, Luan, and
  Sun}]{zhang2017earth}
Meng Zhang, Yang Liu, Huanbo Luan, and Maosong Sun. 2017{\natexlab{b}}.
\newblock Earth mover’s distance minimization for unsupervised bilingual
  lexicon induction.
\newblock In \emph{Proceedings of the 2017 Conference on Empirical Methods in
  Natural Language Processing}, pages 1934--1945.

\bibitem[{Zhang et~al.(2016)Zhang, Gaddy, Barzilay, and
  Jaakkola}]{gaddy2016ten}
Yuan Zhang, David Gaddy, Regina Barzilay, and Tommi Jaakkola. 2016.
\newblock Ten pairs to tag {--} multilingual {POS} tagging via coarse mapping
  between embeddings.
\newblock In \emph{Proceedings of the 2016 Conference of the North {A}merican
  Chapter of the Association for Computational Linguistics: Human Language
  Technologies}, pages 1307--1317.

\end{thebibliography}

\section*{Appendix}

\appendix

\section{Experiment Environment and Settings}
Our experiment is running on a Linux machine with GTX 1080Ti. The version of cuDNN is 7.6.0 and the version of CUDA is 10.1. We also use the PyTorch deep learning framework. The version of PyTorch is 1.6. The average runtime in our experiment is 5 to 10 minutes for one language pair, excluding the time for training word embeddings by fastText. 

Regarding Wikipedia corpora used in our experiments, their download links are: \url{https://dumps.wikimedia.org/${lg}wiki/latest/${lg}wiki-latest-pages-articles.xml.bz2}, \\
where \$\{lg\} should be replaced with the corresponding languages. \$\{lg\} can be set as ``en'', ``es'', ``ar'', ``zh'', ``de'', and ``fr''. We download them on Jan. 2, 2021. 
Regarding WaCKy corpora, we use WaCKy corpora\footnote{\url{https://corpus.leeds.ac.uk/serge/bucc}} provided by BUCC2020 for obtaining the contextual representations, and use the provided word embeddings trained on this corpora.
The BUCC2020 dictionaries are downloaded from the same page. We concat the three training sets with high, mid, low frequency as one training set, and concat the three test sets with high, mid, low frequency as one test set. Because no translations are provided in the test set of BUCC2020, we look up the Muse dictionaries to get the translations. All source words in the test set can find translations in the Muse dictionaries. 

The evaluation script for computing precision@1 (P@1) is: \url{https://github.com/artetxem/vecmap.git/eval_translation.py}.

\section{Parameter Settings}

The parameter size of the spring network is 796202. We use default parameter settings of VecMap and RCSLS. 

\begin{table}[htbp]
\small
\centering
\begin{tabular}{c|c}
\hline
language      & $\lambda$     \\ \hline
EN$\to$ES & 0.11  \\
ES$\to$EN & 0.05 \\
EN$\to$AR & 0.10  \\
AR$\to$EN & 0.30  \\
EN$\to$ZH & 0.12  \\
ZH$\to$EN & 0.10  \\
EN$\to$DE & 0.11 \\
DE$\to$EN & 0.25 \\
EN$\to$FR & 0.13 \\
FR$\to$EN & 0.11 \\
\hline
\end{tabular}
\caption{The optimal hyperparameter $\lambda$ in the unsupervised settings }
\label{tab:hyp}
\end{table}

Regarding the hyperparameter $\lambda$ in the similarity interpolation, we search the optimal value in [0.05-0.3] with step size of 0.01. We found $\lambda=0.1$ is superior on validation sets in all supervised settings. The optimal value of $\lambda$ found by the unsupervised tuning procedure in the unsupervised settings, which is introduced in section 3.2, is shown in Table \ref{tab:hyp}. In the search range, the performances has low variance, and are better than the baselines. 

\section{The Influence of Selecting Encoder Layer in Other Language Pairs}

\begin{figure}[t]
\centering
\subfigure[EN-AR]{
        \includegraphics[width=0.96\linewidth,height=40mm]{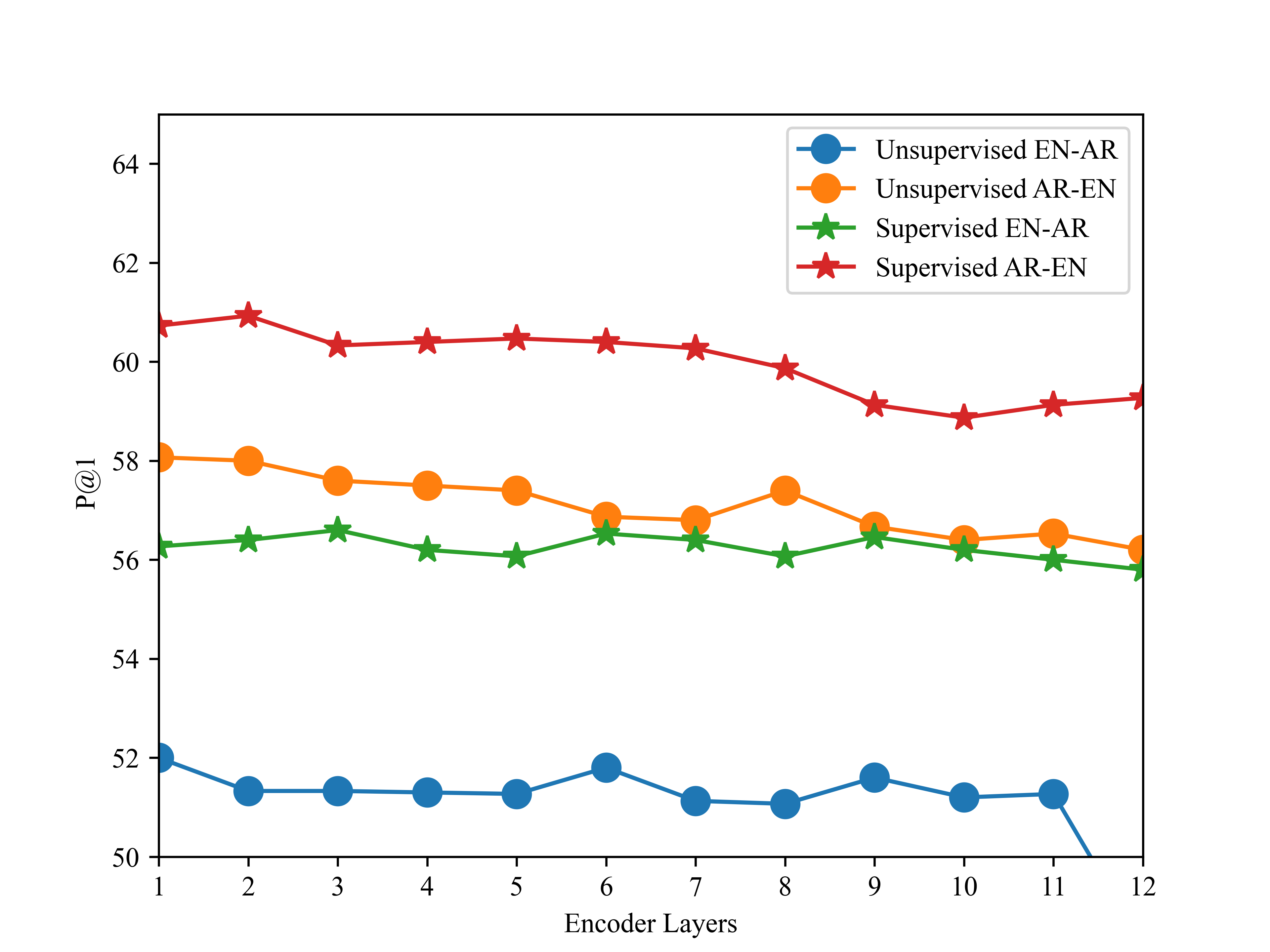}
}
\subfigure[EN-DE]{
        \includegraphics[width=0.96\linewidth,height=40mm]{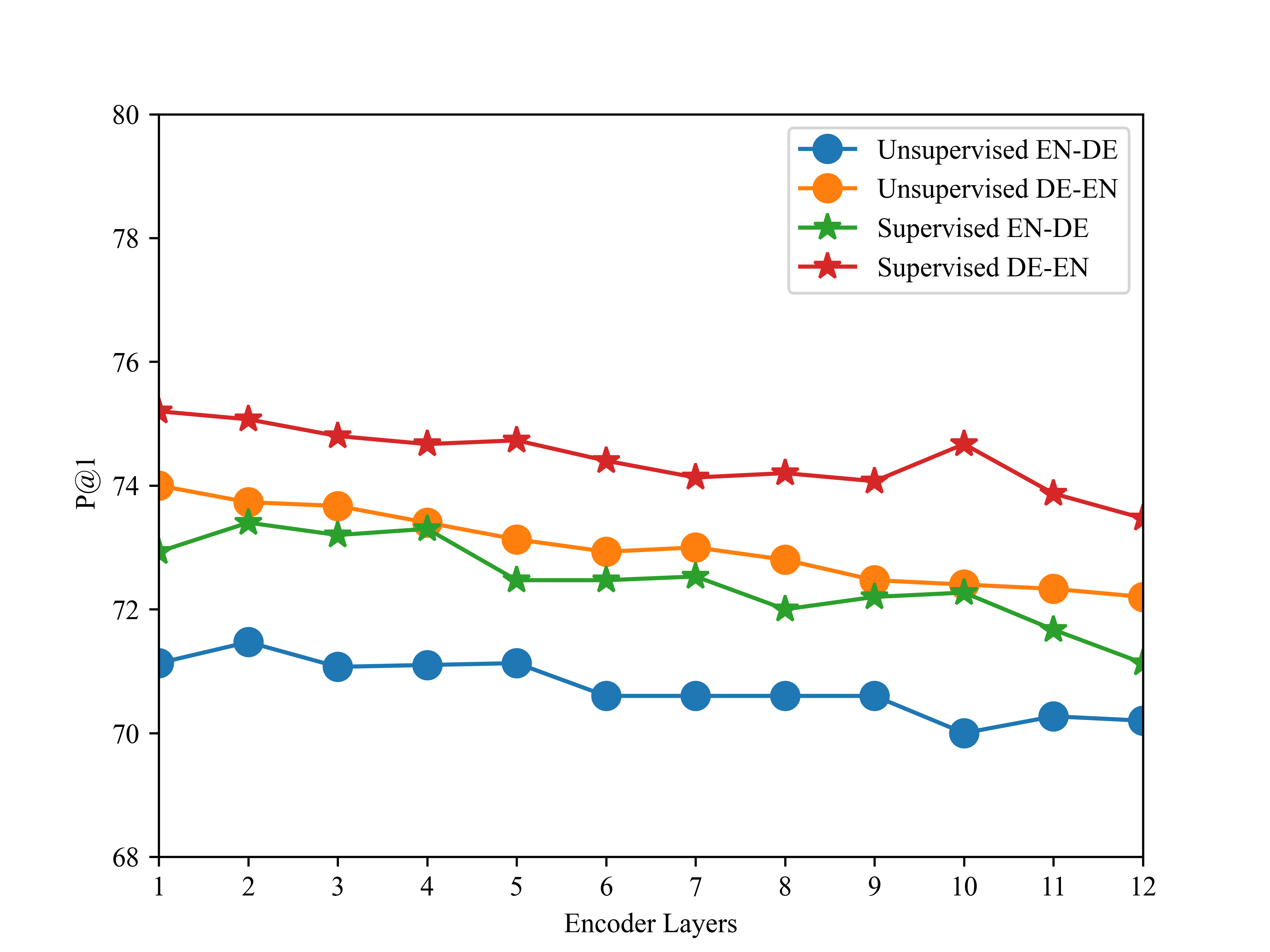}
}
\subfigure[EN-FR]{
        \includegraphics[width=0.96\linewidth,height=40mm]{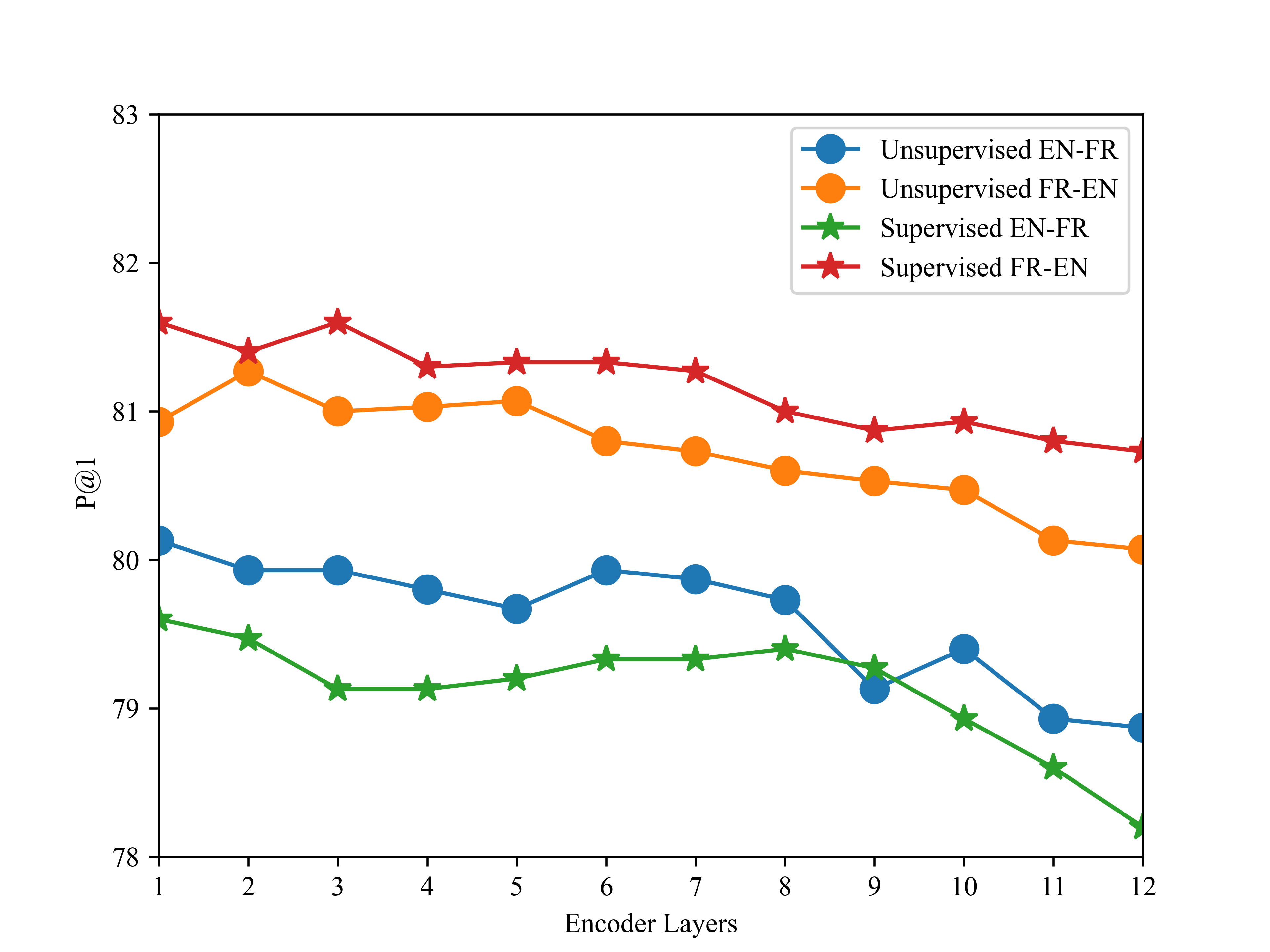}
}
\subfigure[EN-ZH]{
        \includegraphics[width=0.96\linewidth,height=40mm]{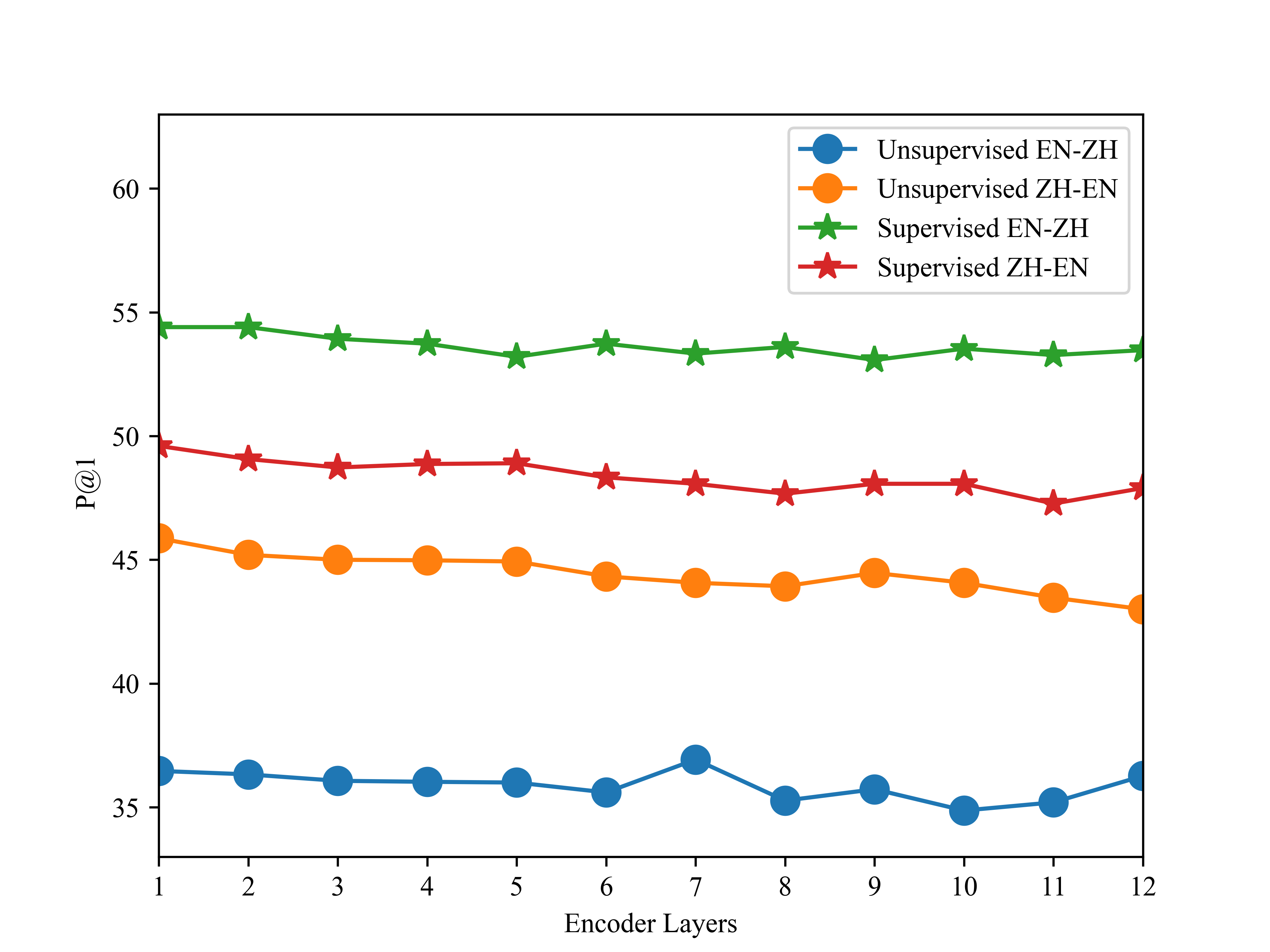}
}
\hspace{0in}
\caption{Performances of selecting different layer of XLM encoder to derive the contextual representations.}
\label{fig:performaceOfLayer}
\end{figure}

We report the influences on EN-AR, EN-ZH, EN-DE, EN-FR in Figure \ref{fig:performaceOfLayer}. It shows that as we select higher layers for deriving the contextual representations, the performances become lower. This trend exists in most language pairs, except that the trend in EN-ZH is not significant.

\section{The Randomness of Contexts in Other Language Pairs}

\begin{figure*}[t]
\flushleft
\subfigure[EN-AR]{
        \includegraphics[width=0.50\linewidth,height=50mm]{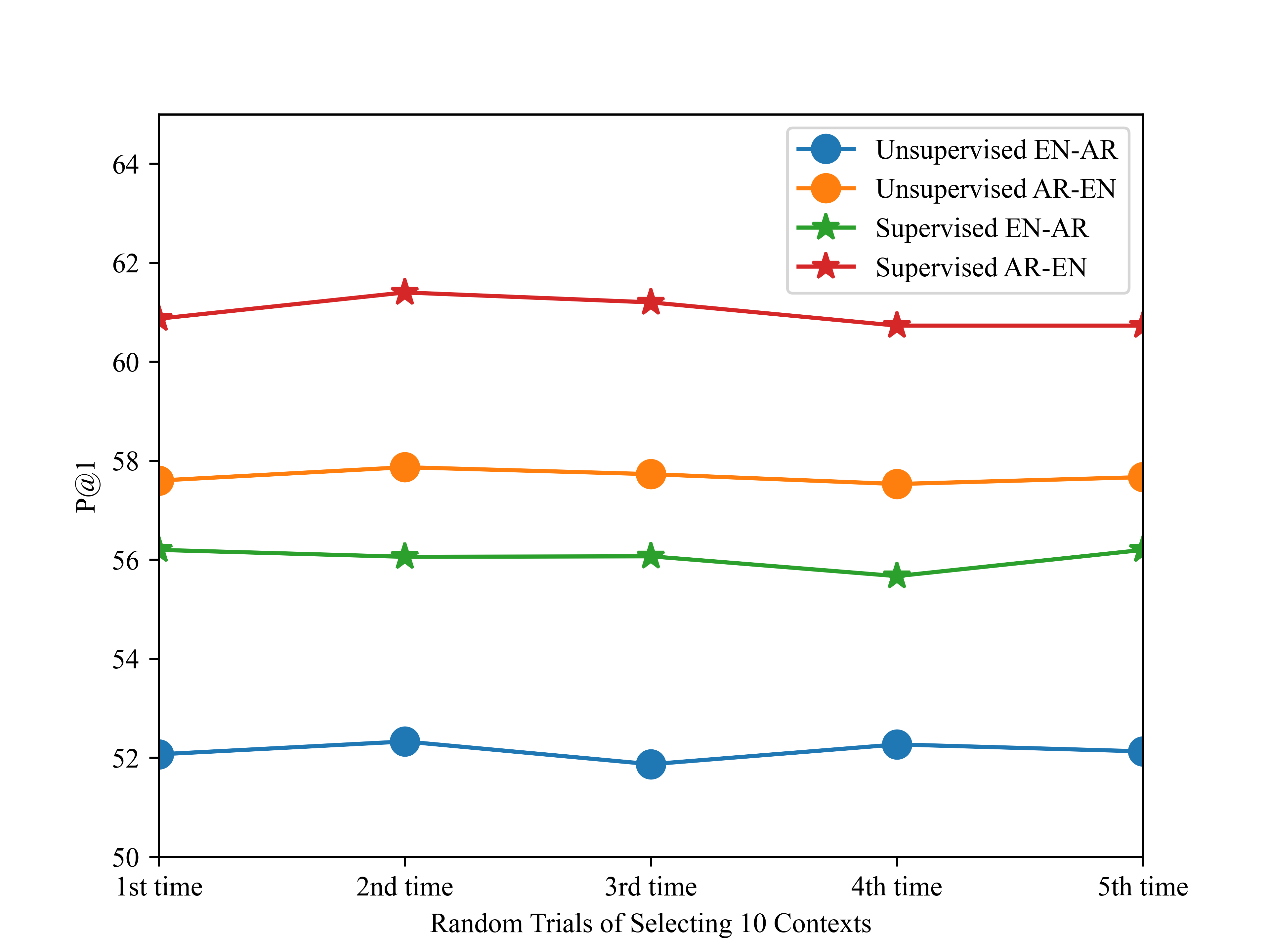}
         \includegraphics[width=0.50\linewidth,height=50mm]{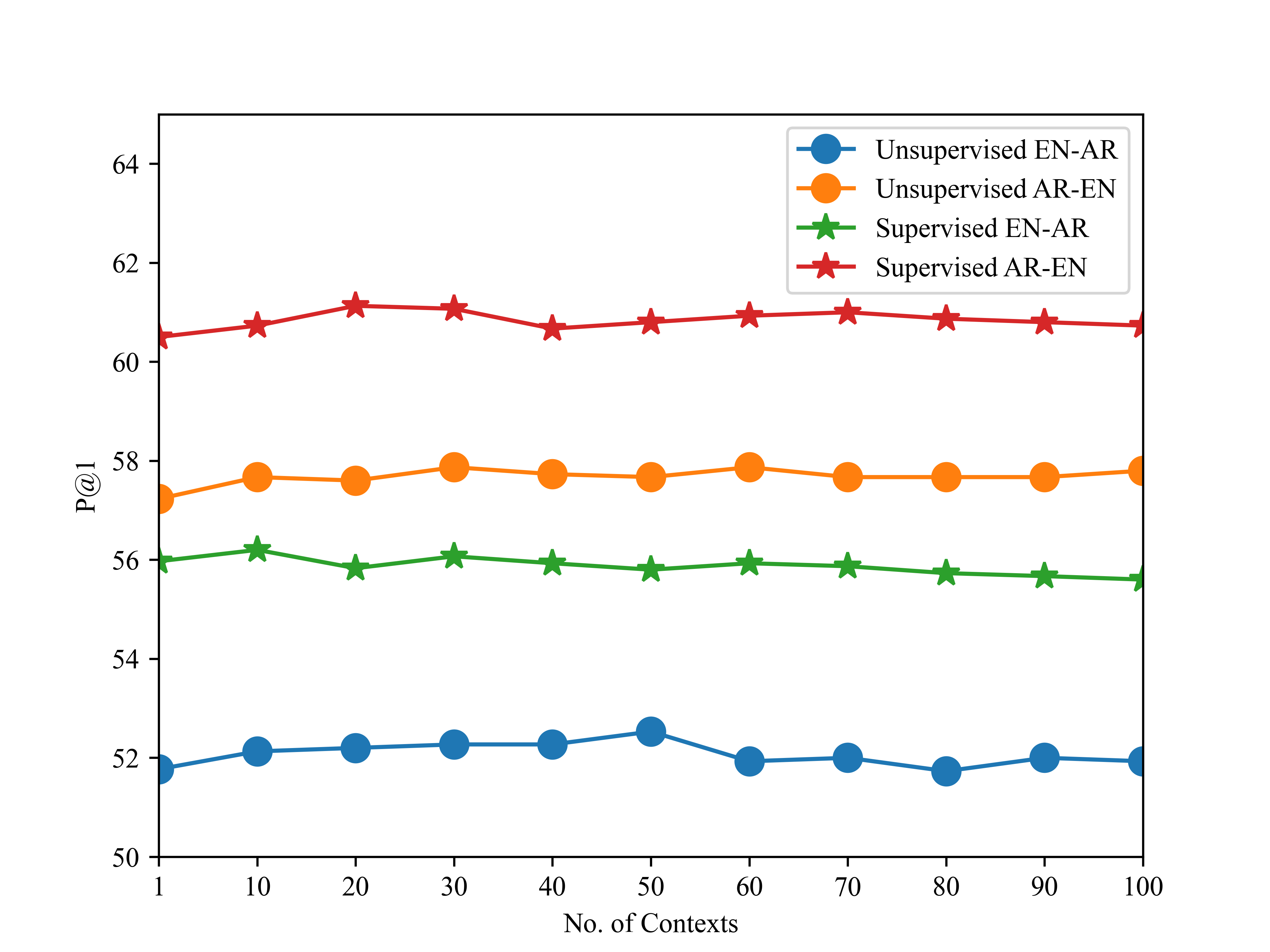}
}
\hspace{0in}
\subfigure[EN-ZH]{
        \includegraphics[width=0.50\linewidth,height=50mm]{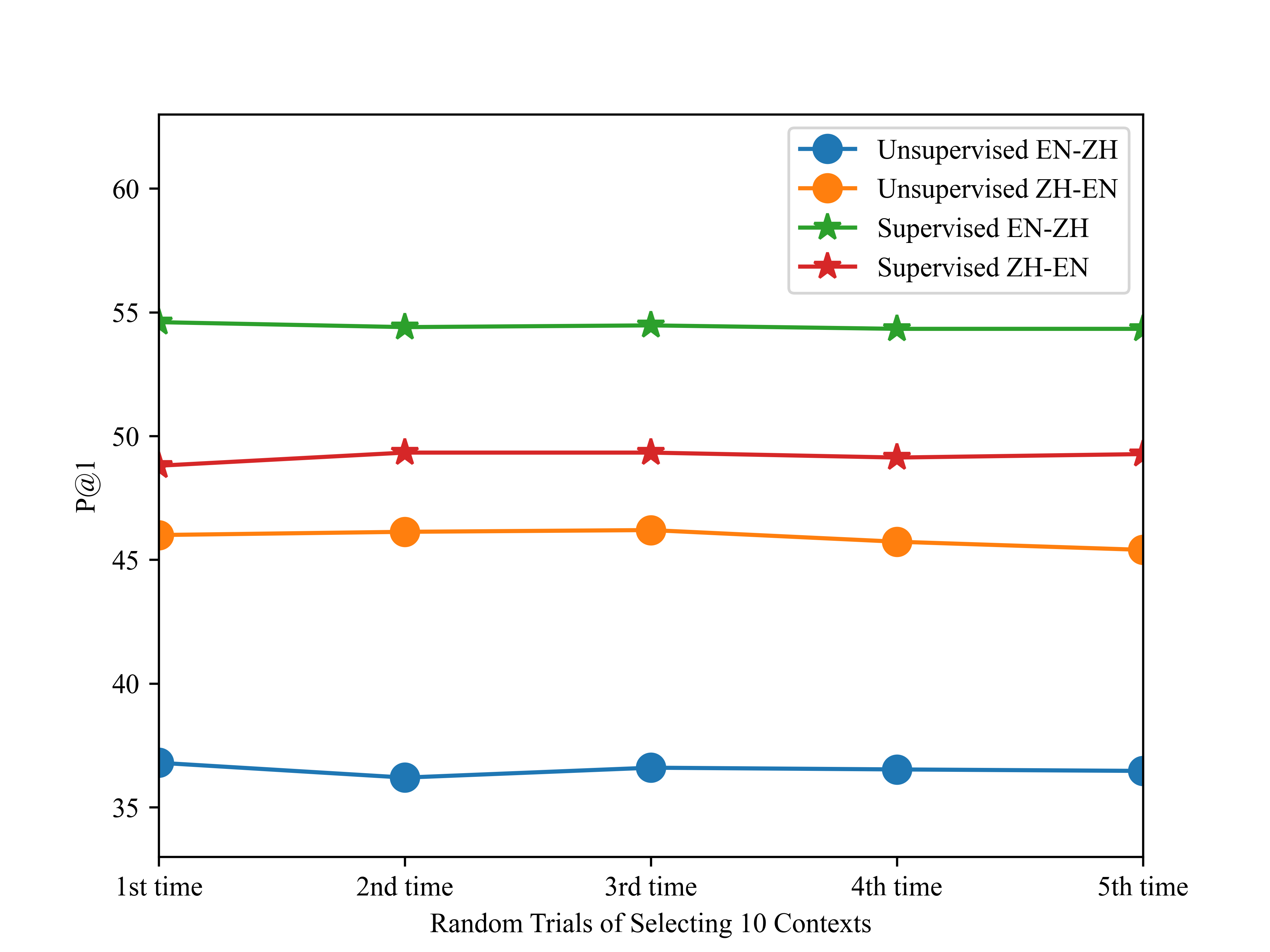}
         \includegraphics[width=0.50\linewidth,height=50mm]{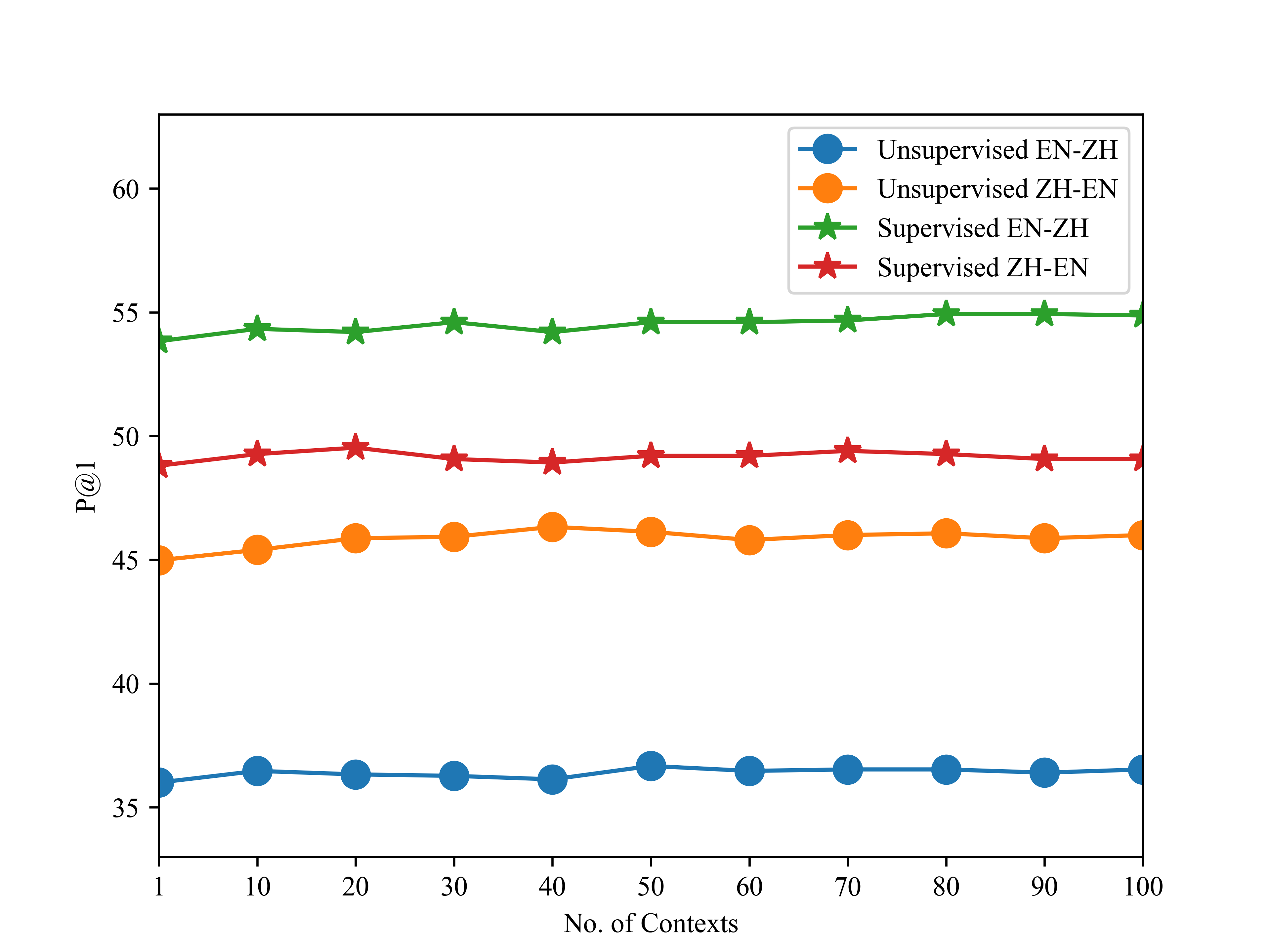}
}
\subfigure[EN-DE]{
        \includegraphics[width=0.50\linewidth,height=50mm]{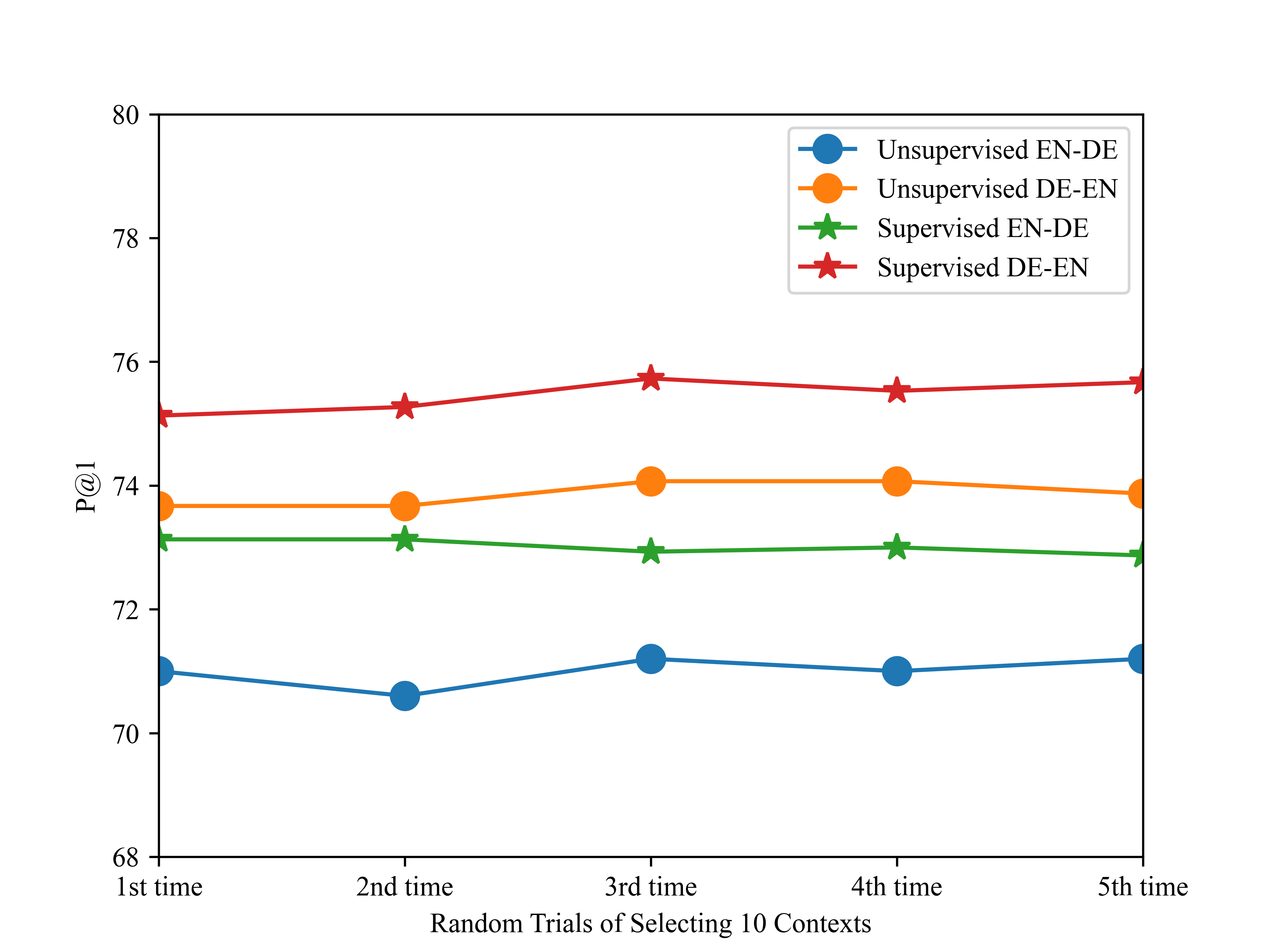}
         \includegraphics[width=0.50\linewidth,height=50mm]{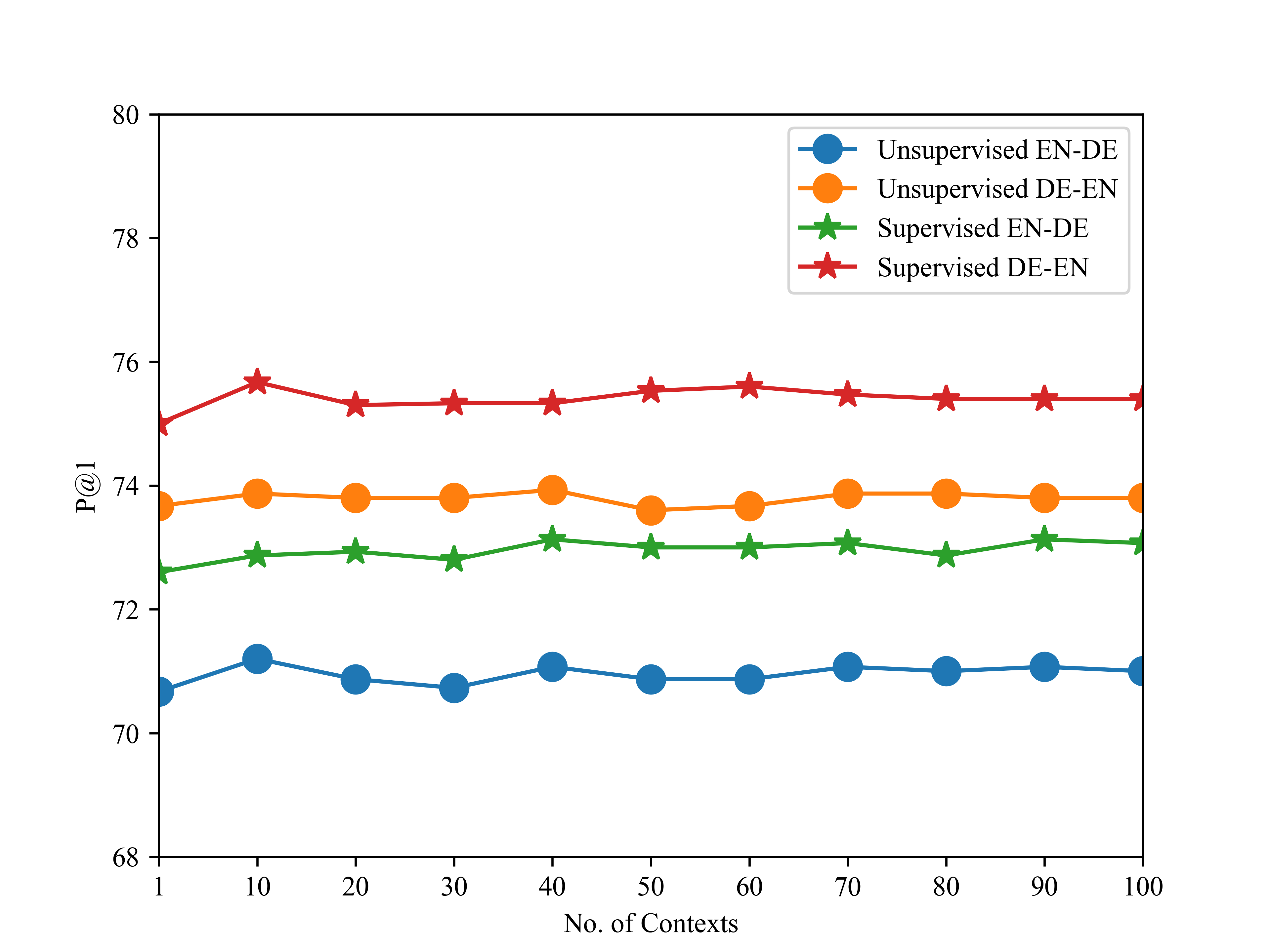}
}
\hspace{0in}
\subfigure[EN-FR]{
        \includegraphics[width=0.50\linewidth,height=50mm]{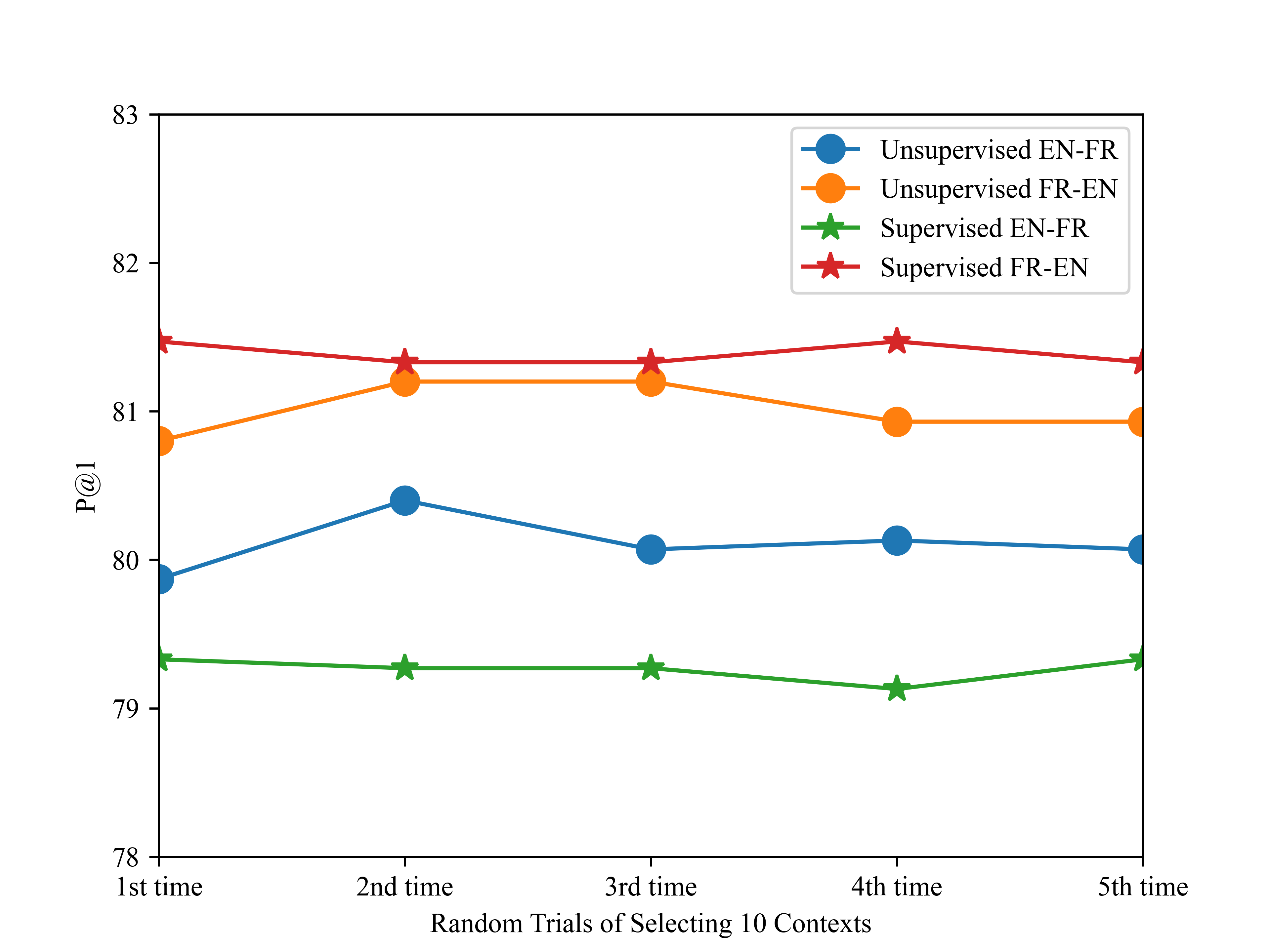}
         \includegraphics[width=0.50\linewidth,height=50mm]{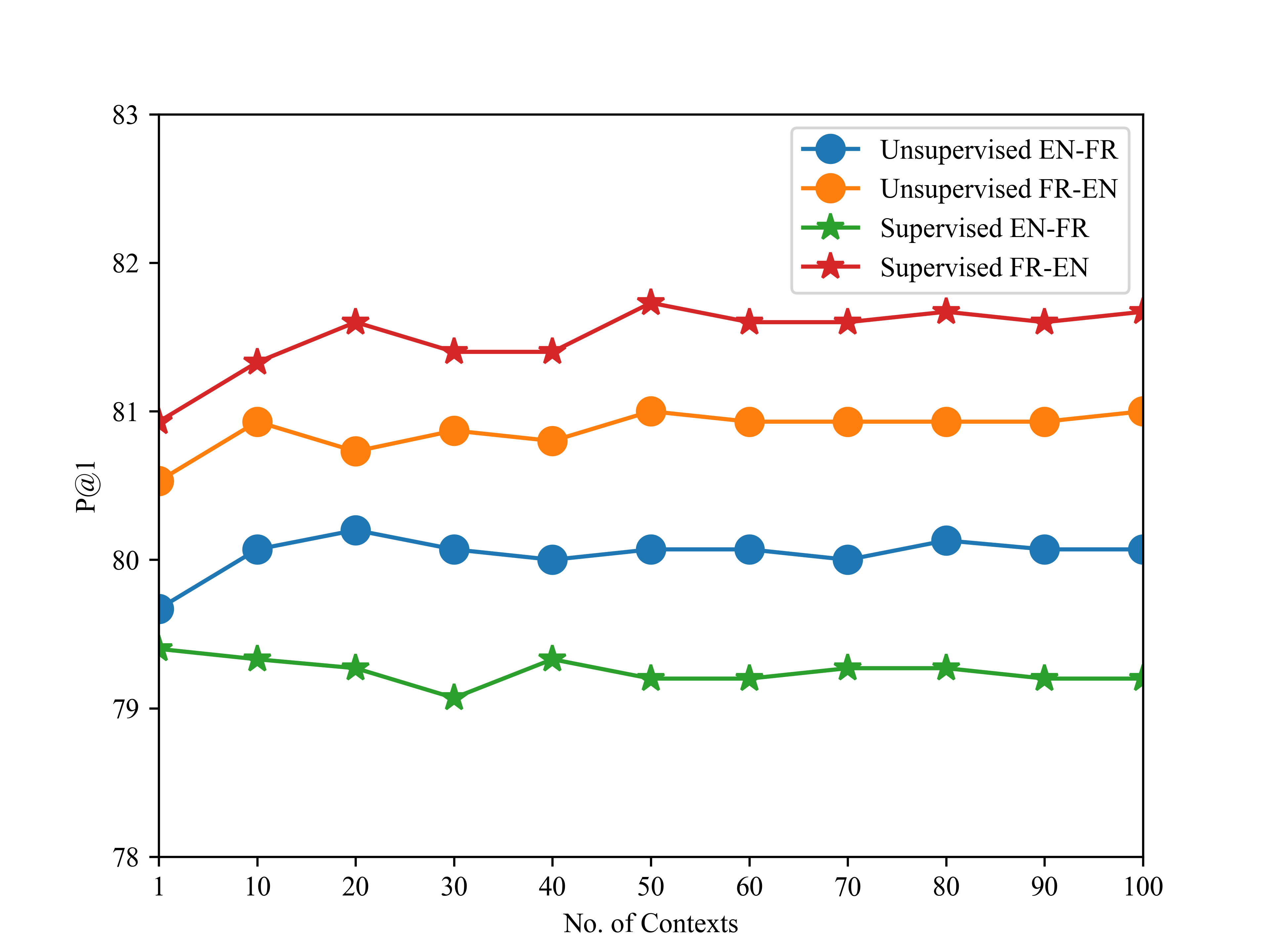}
}

\hspace{0in}
\caption{Performances of different context selection methods on EN-AR, EN-ZH, EN-DE, and EN-FR.}
\label{fig:randomContext1}
\end{figure*}

We report the randomness analyses on EN-AR, EN-ZH, EN-DE, EN-FR in Figure \ref{fig:randomContext1}. It shows that trying 5 times of selecting 10 random sentences for gathering contexts gets stable performances in all language pairs. In most cases, using 1 sentence for computing the contextual representation drags the performance down, which indicates the inadequacy of 1 sentence for gathering contexts. 

\section{Performances on Words with Different Frequencies}

We use WaCKy corpora and the dictionaries provided by BUCC2020\footnote{https://comparable.limsi.fr/bucc2020/bucc2020-task.html} to study the performances on words with different frequencies. The provided dictionaries are divided into groups of high frequency words, mid frequency words, and low frequency words. We test our combination mechanism on these three groups respectively. The results are presented in Table \ref{tbl:unsup_freq} and Table \ref{tbl:sup_freq}. We can see that our combination mechanism is effective on all three groups. 

\begin{table}[htbp]
\scriptsize
\centering
\begin{tabular}{l|c|c|c|c}
 \bottomrule[1.2pt] 
 & EN$\to$DE & DE$\to$EN & EN$\to$FR & FR$\to$EN \\ \hline
\multicolumn{5}{c}{High Frequency}  \\ \hline
VecMap & 67.60 & 75.27 & 79.00 &79.27  \\ \hline
 Unified& 71.27 & 77.40 & 79.67 & 80.80 \\
 Interpolation&  72.53&  79.67& 80.47 & 82.13 \\ \hline
 \multicolumn{5}{c}{Middle Frequency}  \\ \hline
 VecMap& 67.20 &71.47  & 77.73 & 80.27 \\ \hline
 Unified&  67.80	&73.53	&78.33	&81.00\\
 Interpolation&  70.60	&78.13	&81.13	&83.73 \\ \hline
 \multicolumn{5}{c}{Low Frequency}  \\ \hline
 VecMap& 69.53	&74.73	&78.47	&81.87 \\ \hline
 Unified&  70.00	&75.53	&79.27	&82.67 \\
 Interpolation&  77.53	&82.47	&87.00	&87.07 \\ 
  \bottomrule[1.2pt] 

\end{tabular} 
 \caption{Performances on words with different frequencies in the unsupervised setting}
 \label{tbl:unsup_freq}
\end{table}

\begin{table}[htbp]
\scriptsize
\centering
\begin{tabular}{l|c|c|c|c}
 \bottomrule[1.2pt] 
 & EN$\to$DE & DE$\to$EN & EN$\to$FR & FR$\to$EN \\ \hline
\multicolumn{5}{c}{High Frequency}  \\ \hline
VecMap & 67.87	&76.13&	78.27	&79.73 \\ \hline
 Unified& 70.40	&78.07&	78.4	&81.4 \\
 Interpolation& 71.00	&78.90	&78.90&	82.13 \\ \hline
 \multicolumn{5}{c}{Middle Frequency}  \\ \hline
 VecMap& 68.87&	71.8	&78.47&	80.87 \\ \hline
 Unified&  69.93&	75.20&	79.53	&82.20\\
 Interpolation&  71.20&	76.33&	80.07&	83.73\\ \hline
 \multicolumn{5}{c}{Low Frequency}  \\ \hline
 VecMap& 70.60	&76.13	&79.67&	82.93 \\ \hline
 Unified&72.67&	78.27&	81.20	&84.20 \\
 Interpolation &74.13	&79.97&	82.30	&84.73\\ 
  \bottomrule[1.2pt] 
\end{tabular} 
 \caption{Performances on words with different frequencies in the supervised setting}
 \label{tbl:sup_freq}
\end{table}

\end{document}